\documentclass[10pt,twocolumn,letterpaper]{article}

\pdfoutput=1

\usepackage{iccv}
\usepackage{times}
\usepackage{epsfig}
\usepackage{graphicx}
\usepackage{amsmath}
\usepackage{amssymb}

\usepackage{url}
\usepackage{multirow}
\usepackage{subfigure}
\usepackage{diagbox}
\usepackage{makecell}
\usepackage{wrapfig}
\usepackage{bm}
\usepackage{xcolor}

\usepackage[pagebackref=true,breaklinks=true,letterpaper=true,colorlinks,bookmarks=false]{hyperref}

\iccvfinalcopy % *** Uncomment this line for the final submission

\def\iccvPaperID{3090} % *** Enter the ICCV Paper ID here

\newcommand\blfootnote[1]{% 
\begingroup 
\renewcommand\thefootnote{}\footnote{#1}% 
\addtocounter{footnote}{-1}% 
\endgroup 
}

\begin{document}

%%%%%%%%% TITLE
\title{IF-Defense: 3D Adversarial Point Cloud Defense via Implicit Function based Restoration}

\author{Ziyi Wu$^{1,*}$\qquad 
Yueqi Duan$^{2,*,\dagger}$\qquad 
He Wang$^{2}$\qquad 
Qingnan Fan$^{2}$\qquad 
Leonidas J. Guibas$^{2}$\\
$^{1}$Tsinghua University\qquad 
$^{2}$Stanford University
}

\maketitle

%%%%%%%%% ABSTRACT
\begin{abstract}
Point cloud is an important 3D data representation widely used in many essential applications. Leveraging deep neural networks, recent works have shown great success in processing 3D point clouds. 
However, those deep neural networks are vulnerable to various 3D adversarial attacks, which can be summarized as two primary types: \emph{point perturbation} that affects local point distribution, and \emph{surface distortion} that causes dramatic changes in geometry. 
In this paper, we simultaneously address both the aforementioned attacks by learning to restore the clean point clouds from the attacked ones. More specifically, we propose an IF-Defense framework to directly optimize the coordinates of input points with geometry-aware and distribution-aware constraints. The former aims to recover the surface of point cloud through implicit function, while the latter encourages evenly-distributed points. 
Our experimental results show that IF-Defense achieves the state-of-the-art defense performance against existing 3D adversarial attacks on PointNet, PointNet++, DGCNN, PointConv and RS-CNN. For example, compared with previous methods, IF-Defense presents 20.02\% improvement in classification accuracy against salient point dropping attack and 16.29\% against LG-GAN attack on PointNet. The code is available at \url{https://github.com/Wuziyi616/IF-Defense}.
\blfootnote{* Equal contribution}
\blfootnote{$\dagger$ Corresponding author: duanyq19@stanford.edu}
\end{abstract}

%%%%%%%%% BODY TEXT
\section{Introduction}

Recent years have witnessed a growing popularity of various 3D sensors such as LiDAR and Kinect in self-driving cars, robotics and AR/VR applications. As the direct outputs of these sensors, point cloud has drawn increasing attention. Point cloud is a compact and expressive 3D representation, which represents a shape using a set of unordered points and can capture arbitrary complex geometry. However, the irregular data format makes point clouds hard to be directly processed by deep neural networks (DNNs). To address this, PointNet \cite{qi2017pointnet} first uses multi-layer perceptrons (MLPs) to extract point-wise features and aggregate them with max-pooling. Since then, a number of studies \cite{qi2017pointnet++, wang2019dynamic, wu2019pointconv, liu2019relation} have been conducted to design 3D DNNs for point clouds and achieve tremendous progress.

\begin{figure*}[t]
    \centering
    \vspace{-16pt}
    \subfigure[Clean Point Cloud]{
        \includegraphics[width=0.3\textwidth]{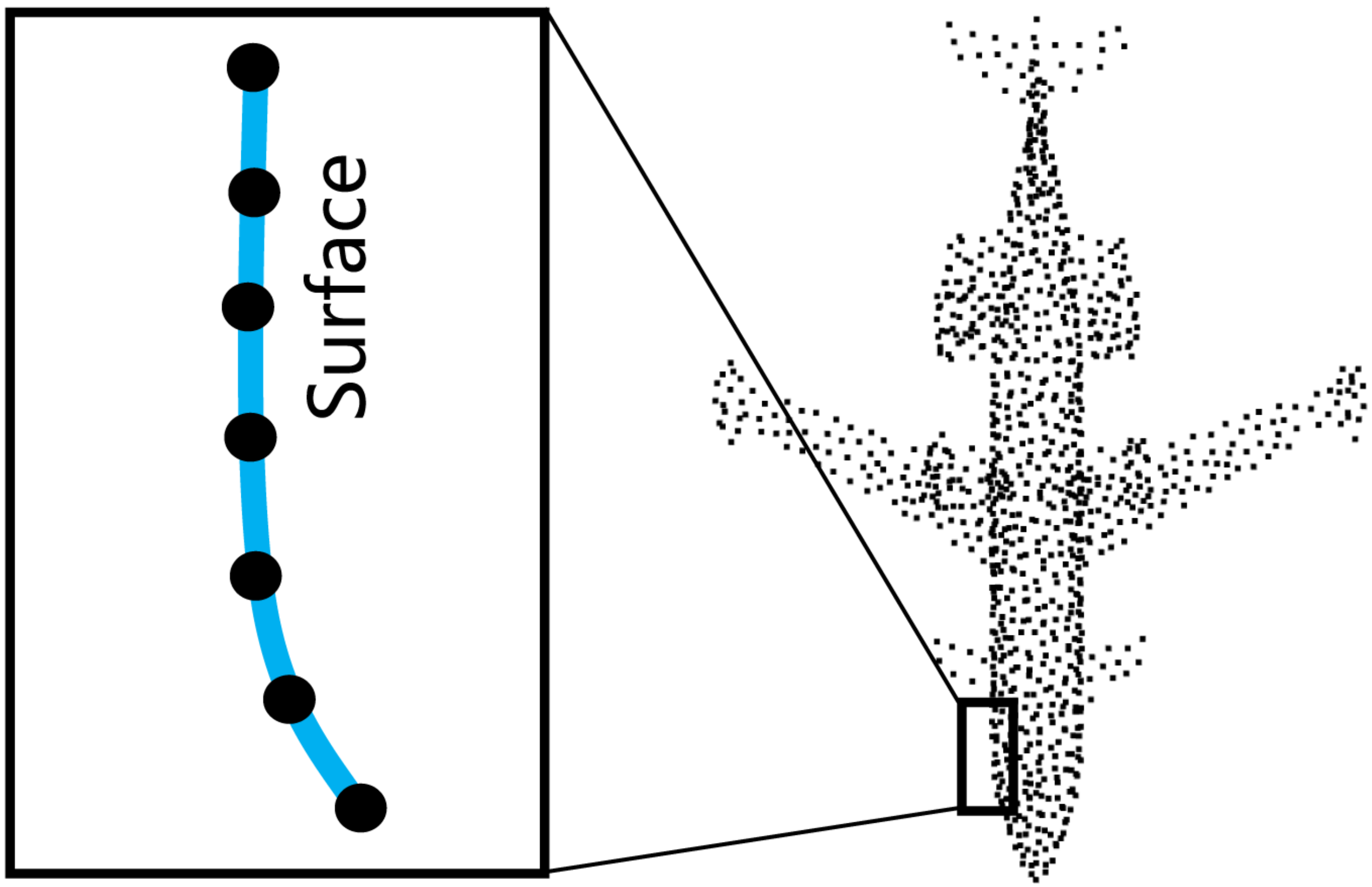}
    }
    \quad%\hspace{0.2cm}
    \subfigure[Out-of-surface Perturbation]{
        \includegraphics[width=0.3\textwidth]{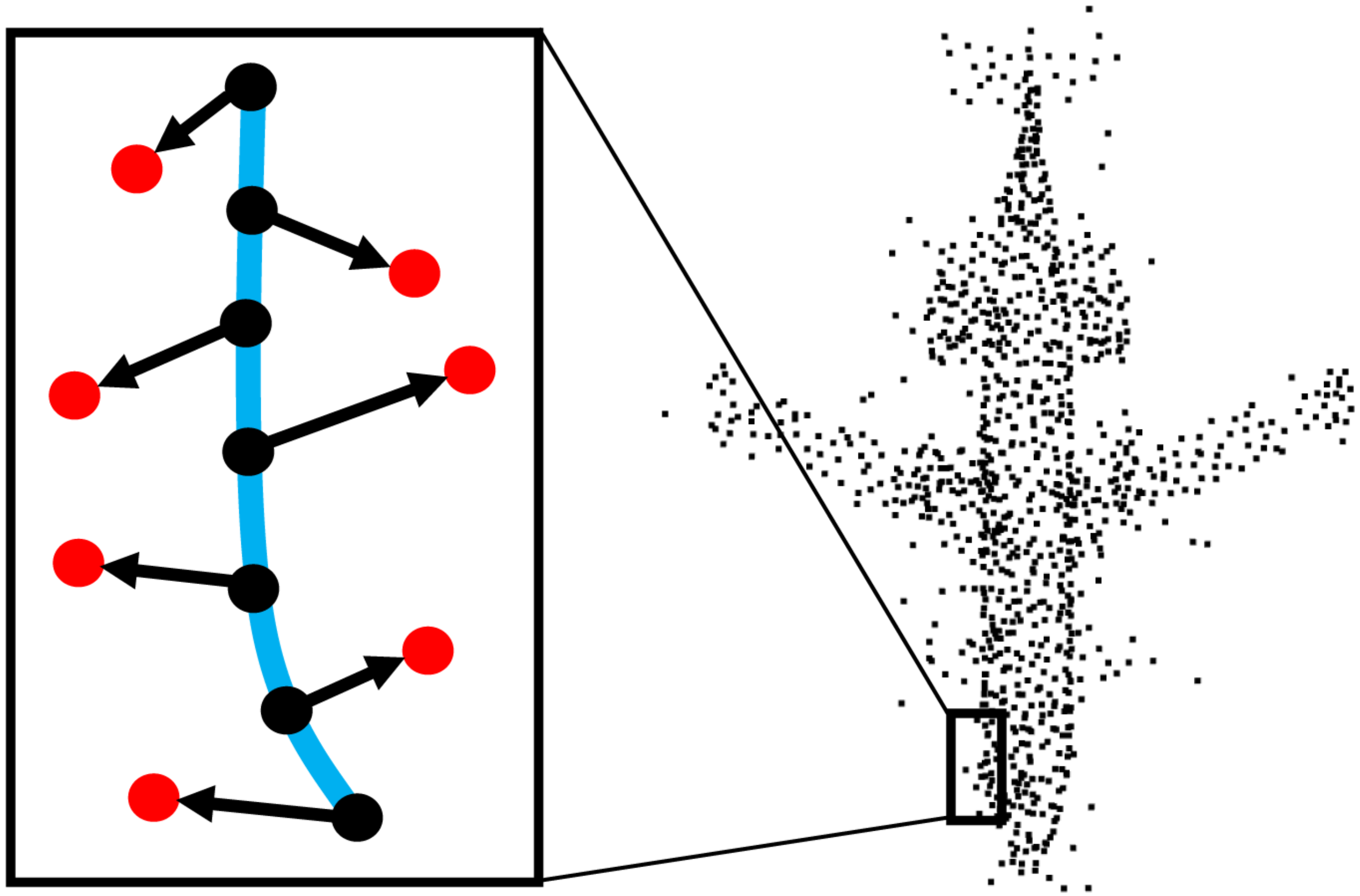}
    }
    \quad%\hspace{0.2cm}
    \subfigure[On-surface Perturbation]{
        \includegraphics[width=0.3\textwidth]{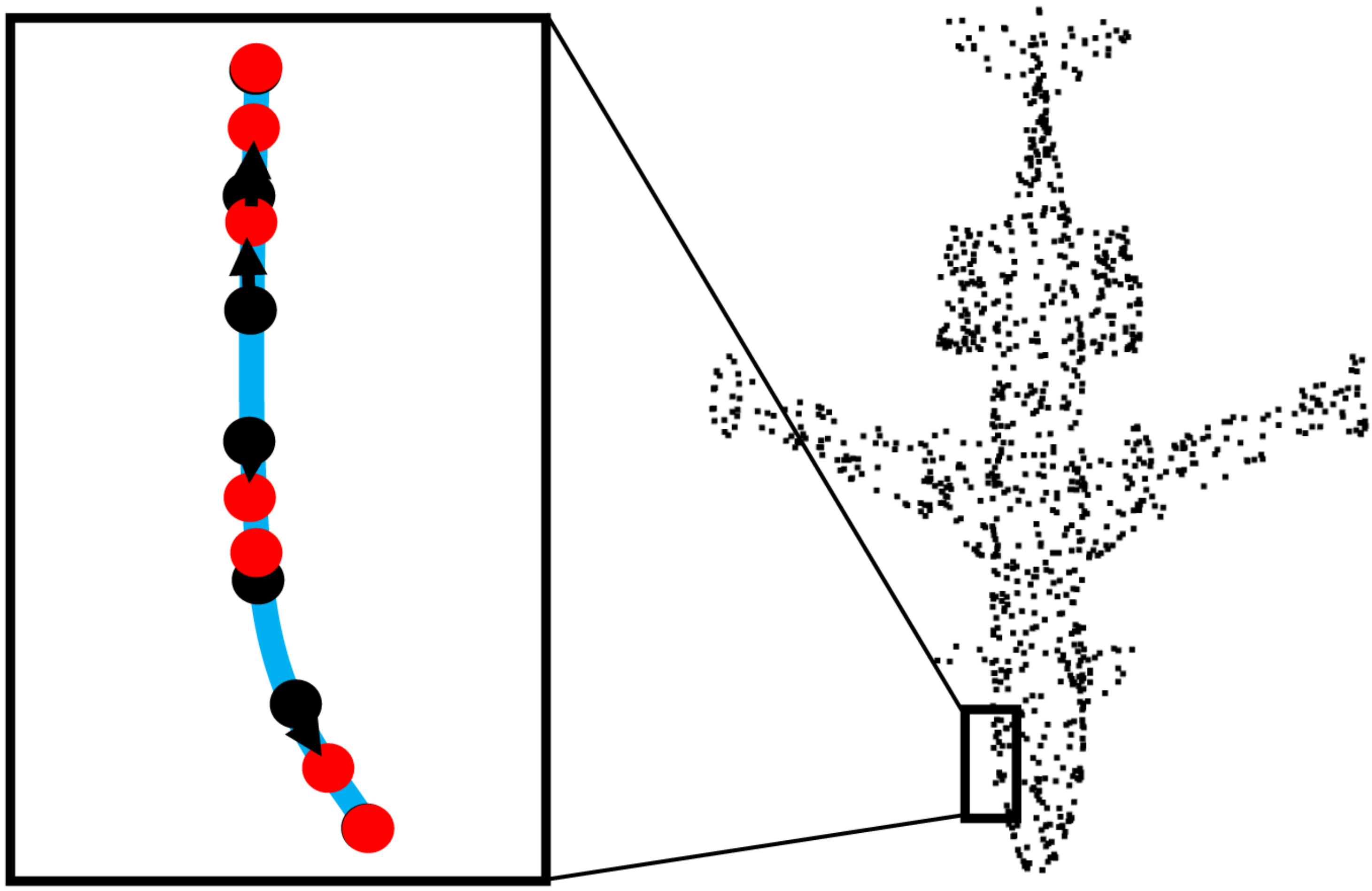}
    }
    \quad%\hspace{0.2cm}
    \subfigure[Local Part Removal]{
        \includegraphics[width=0.3\textwidth]{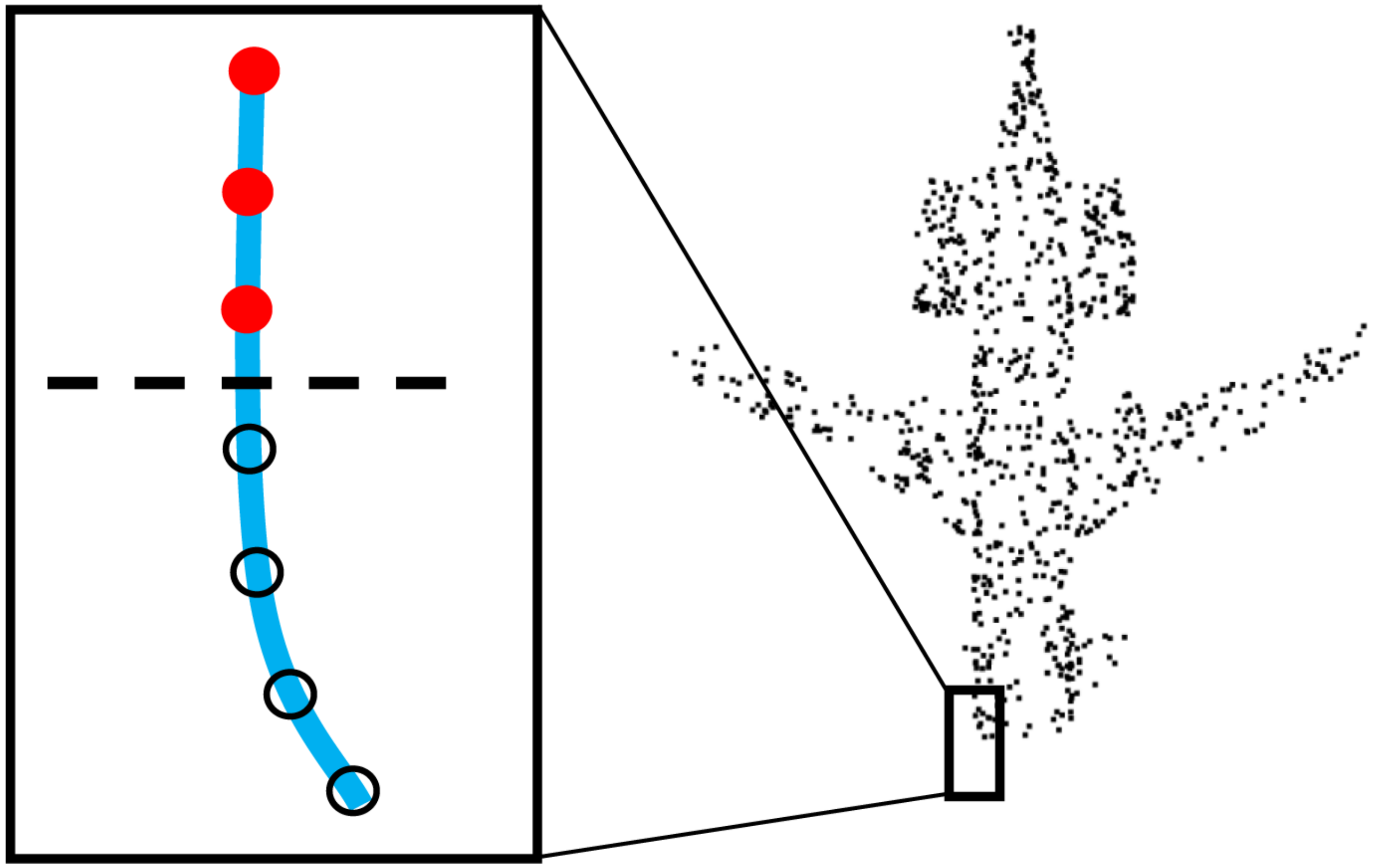}
    }
    \quad%\hspace{0.2cm}
    \subfigure[Geometric Deformation]{
        \includegraphics[width=0.3\textwidth]{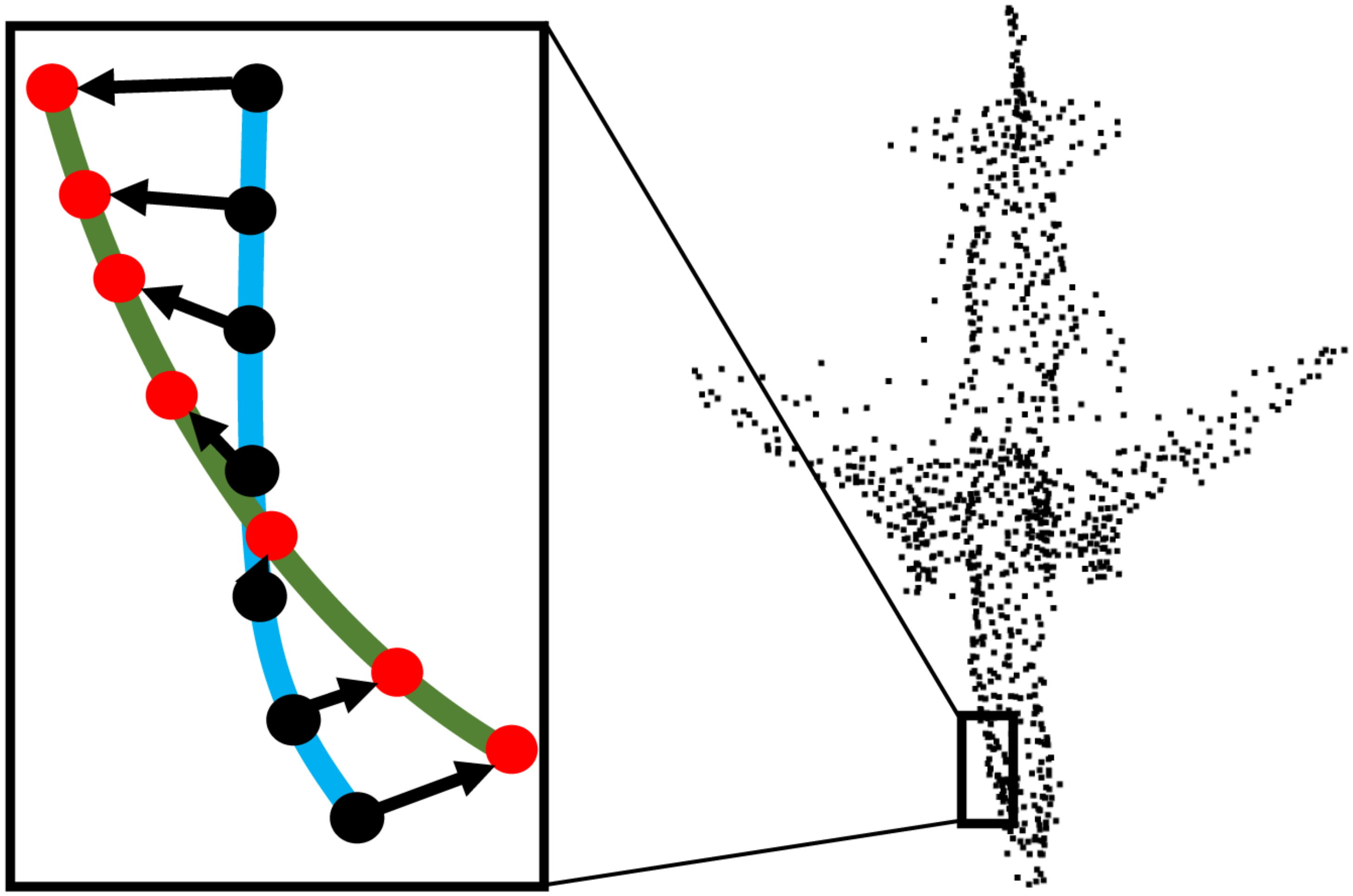}
    }
    %\vspace{-3mm}
    \caption{The key effects of 3D adversarial attacks on point cloud summarized from existing works. We show (a) a clean point cloud, (b)(c) point perturbation, and (d)(e) surface distortion. In each subfigure, we show an entire shape and a local illustration. The blue curve is the object surface, the black points are clean points and the red points are attacked points.}
    \label{attack_effects}
    \vspace{-12pt}
    %\vspace{-5mm}
\end{figure*}

One limitation of DNNs is that they are vulnerable to adversarial attacks. By adding imperceptible perturbations to clean data, the generated adversarial examples can mislead victim models with high confidence. While numerous algorithms have been proposed in 2D attack and defense \cite{goodfellow2014explaining, carlini2017towards, xie2017mitigating, papernot2016distillation, moosavi2016deepfool, athalye2018synthesizing, moosavi2017universal}, only a little attention is paid to its 3D counterparts \cite{xiang2019generating, zhou2019dup, zheng2019pointcloud}. They show that point cloud networks such as PointNet \cite{qi2017pointnet} and PointNet++ \cite{qi2017pointnet++} are also sensitive to adversarial examples. 
Besides, recent works \cite{cao2019adversarial, tu2020physically} have conducted physically realizable point cloud attacks on autonomous driving and robotics tasks in real-world scenarios, posing severe threat to these safety-critical applications. 
By carefully examining existing 3D adversarial attack methods, we summarize their attack effects into two aspects as shown in Figure \ref{attack_effects}:

\begin{enumerate}
\renewcommand{\labelenumi}{\theenumi) }
    \item \emph{Point perturbation} changes the local geometry and point-wise sampling pattern, which moves the points either out of the surface to become noises or along the surface to change point distributions. 
    This effect is similar to 2D adversarial attack, which adds noise over each pixel within a given budget to fool the classifier.

    \item \emph{Surface distortion} aims to modify the geometry of the point cloud more dramatically by either removing local parts or deforming the shape of the point cloud. 
    In general, surface distortion is difficult to defend due to the significant change of the geometry, yet is also more perceptible by humans.
\end{enumerate}

% attack effects visualization

While some methods have been proposed in recent years for 3D adversarial defense \cite{zhou2019dup, dong2020self}, they fail to simultaneously address both the two aspects. For example, DUP-Net \cite{zhou2019dup} uses a statistical outlier removal (SOR) pre-processor to address out-of-surface point perturbations, followed by an up-sampling network to generate denser point clouds. However, it cannot well recover the point distribution and restore the distorted surface. Gather-vector guidance (GvG) method \cite{dong2020self} learns to ignore noisy local features, which fails to defend the attacks by local part removal. As a result, these methods fail to protect the victim models from all the attacks, especially the latest ones, such as salient point dropping \cite{zheng2019pointcloud}, LG-GAN \cite{zhou2020lg} and AdvPC \cite{hamdi2020advpc}.

\begin{figure*}[tb]
	\centering
	\includegraphics[width=\textwidth]{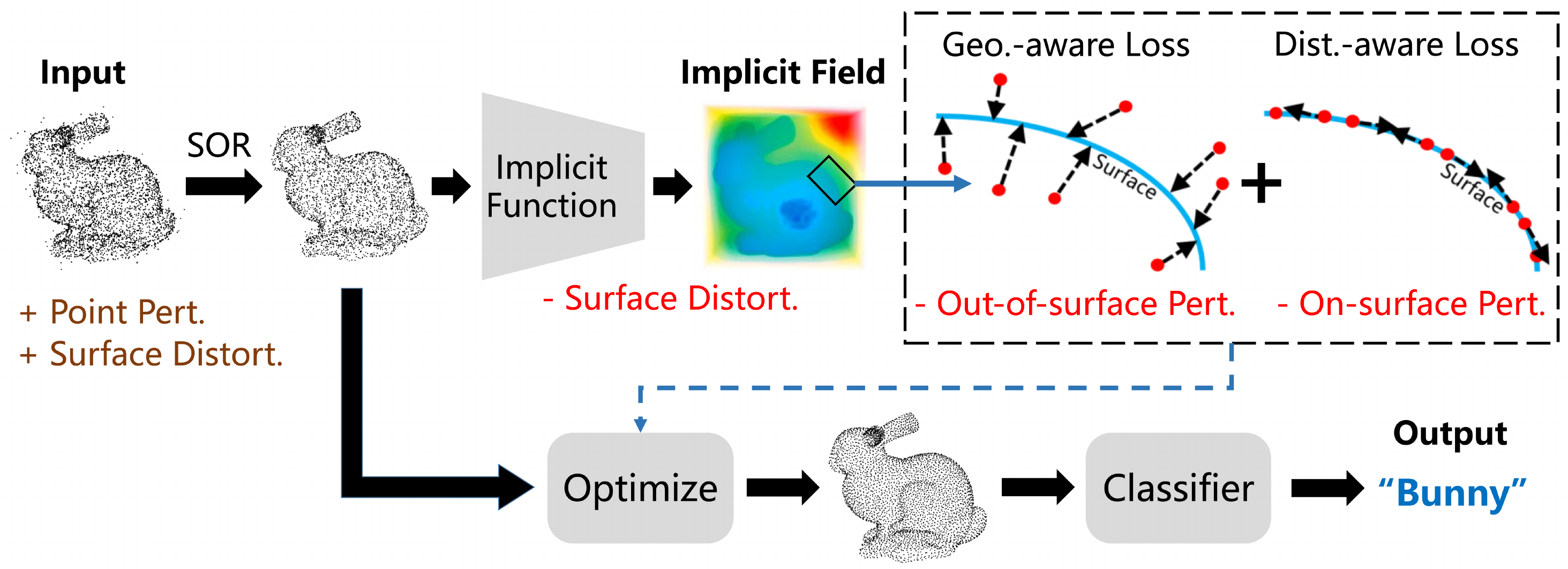}
	\caption{The pipeline of our IF-Defense method. We first pre-process the input point cloud by SOR, and then we learn the point coordinates of the restored point cloud via implicit function based optimization guided by two losses. Finally, we send the restored point cloud to the classifier. \textit{Pert.} and \textit{Distort.} indicate Perturbation and Distortion, while \textit{Geo.} and \textit{Dist.} stand for Geometry and Distribution.}
	\label{method_pipeline}
	\vspace{-10pt}
\end{figure*}

In this paper, we propose a 3D adversarial point cloud defense algorithm named IF-Defense by learning to restore the clean point clouds from the attacked ones, which is more universal and simultaneously addresses both the attack effects. 
Figure \ref{method_pipeline} shows the pipeline of IF-Defense. We first employ SOR to pre-process the input point cloud following the existing work \cite{zhou2019dup}. Then, we directly optimize the coordinates of input points under the geometry-aware and distribution-aware constraints. 
The geometry-aware loss aims to remove out-of-surface geometric changes, such as Figure \ref{attack_effects}(b)(d)(e). Inspired by the recent success in deep implicit functions which reconstruct accurate surfaces even under partial observations \cite{park2019deepsdf, duan2020curriculum, mescheder2019occupancy, peng2020convolutional, chen2019learning}, we train an implicit function network on clean point clouds to estimate the object surfaces. The predicted surface is locally smooth due to the continuity of the output space of implicit functions \cite{mescheder2019occupancy, park2019deepsdf}, which relieves the effects of outliers. 
The distribution-aware loss aims to distribute points evenly and eliminate the on-surface point perturbation, as illustrated in Figure \ref{attack_effects}(c). We maximize the distance between each point and its $k$-nearest neighbors to encourage uniform point distribution. 
Experimental results show that IF-Defense consistently outperforms existing defense methods against various 3D adversarial attacks for PointNet, PointNet++, DGCNN, PointConv and RS-CNN.

\section{Related Works}

\textbf{Deep learning on point clouds.} The pioneering work PointNet \cite{qi2017pointnet} is the first deep learning algorithm that operates directly on 3D point clouds. After that, PointNet++ \cite{qi2017pointnet++} further improves the performance of PointNet by exploiting local information. Another representative work is Dynamic Graph CNN (DGCNN) \cite{wang2019dynamic}, which constructs $k$NN graphs and applies EdgeConv to capture local geometric structures. In recent years, there are more and more convolution based methods proposed in the literature \cite{wu2019pointconv, thomas2019kpconv, hermosilla2018monte, liu2019relation}, which run convolutions across neighboring points using a predicted kernel weight. Though these point cloud networks have achieved promising results, they are vulnerable to adversarial attacks and require defense methods to improve the robustness. %To evaluate the effectiveness of our defense, we conduct experiments on four typical networks: PointNet, PointNet++, DGCNN and PointConv.

% pipeline image

\textbf{3D adversarial attack.} Existing 3D adversarial attack methods can be roughly divided into three classes: optimization based methods, gradient based methods and generation based methods. For optimization based methods, \cite{xiang2019generating} first proposed to generate adversarial point clouds using C\&W attack framework \cite{carlini2017towards} by point perturbation and adding. 
In contrast, \cite{tsai2020robust} proposed to add a $k$NN distance constraint, a clipping and a projection operation to generate adversarial point clouds that are resistant to defense. Besides, \cite{hamdi2020advpc} proposed AdvPC by utilizing a point cloud auto-encoder (AE) to improve the transferability of adversarial examples. Because of the limited budget, these attacks mainly introduce point perturbations. 
For gradient based methods, \cite{liu2019extending} extended the fast/iterative gradient method to perturb the point coordinates. Additionally, \cite{zheng2019pointcloud} developed a point dropping attack by constructing a gradient based saliency map, which would remove important local parts. 
LG-GAN \cite{zhou2020lg} is a generation based 3D attack method, which leverages GANs \cite{goodfellow2014generative} to generate adversarial point clouds guided by the input target labels. %It is faster compared to optimization and gradient based methods, yet it usually causes large deformation in the generated adversarial examples.
We summarize the correspondence between existing 3D attacks and the attack effects in Table \ref{attacks_corr_effects}.

% correspondence between attack and effects
\begin{table}[tb]
    %\vspace{-0.3cm}
    %\scriptsize
    %\footnotesize
    \caption{Correspondence between existing 3D attacks and attack effects. Out Pert., On Pert., LPR and GD stand for out-of-surface perturbation, on-surface perturbation, local part removal and geometric deformation, respectively. In the table, $\checkmark$ indicates the main effects of an attack while $\triangle$ shows the less significant ones.}
    \label{attacks_corr_effects}
	\centering
	\vspace{5pt}
\renewcommand\arraystretch{1.2}
\begin{tabular}{|c|p{1.3cm}<{\centering}|p{1.2cm}<{\centering}|p{1.1cm}<{\centering}|p{1.1cm}<{\centering}|}
    \hline
    \textbf{Attacks} & Out Pert. & On Pert. & LPR & GD\\
    \hline
    \textbf{Perturb} & $\checkmark$ & $\triangle$ & & \\
    \textbf{Add} & $\checkmark$ & $\triangle$ & & \\
    \textbf{$k$NN} & $\triangle$ & $\checkmark$ & & \\
    \textbf{AdvPC} & $\triangle$ & $\checkmark$ & & \\
    \textbf{Drop} & & $\triangle$ & $\checkmark$ & \\
    \textbf{LG-GAN} & $\triangle$ & $\checkmark$ & & $\checkmark$\\
    \hline
\end{tabular}
    \vspace{-15pt}
\end{table}

%\subsection{Existing 3D Adversarial Defense Methods}

\textbf{3D adversarial defense.} 
\cite{liu2019extending} employed adversarial training to improve the robustness of models by training on both clean and adversarial point clouds. \cite{yang2019adversarial} proposed Gaussian noising and point quantization, which are adopted from 2D defense. They also introduced a Simple Random Sampling (SRS) method which samples a subset of points from the input point cloud. Recently, \cite{zhou2019dup} proposed a Statistical Outlier Removal (SOR) method that removes points with a large $k$NN distance. They also proposed DUP-Net, which is a combination of SOR and a point cloud up-sampling network PU-Net \cite{yu2018pu}. The non-differentiability of SOR also improves its robustness. Instead of designing a pre-processing module to recover adversarial examples, \cite{dong2020self} leveraged the intrinsic properties of point clouds and develop a variant of PointNet++ \cite{qi2017pointnet++} that can identify and discard adversarial local parts of an input. Although these defenses are effective against simple attacks \cite{xiang2019generating}, their performance against more complex methods \cite{tsai2020robust, zhou2020lg} is relatively poor, which is because they fail to simultaneously address the aforementioned two attack effects.

\textbf{Implicit representation.} Different from the voxel-based, mesh-based and point-based methods that explicitly represent shape surface, implicit functions learn a continuous field and represent surface as the zeroth level-set. More recently, deep learning based methods use DNNs to approximate the occupancy field \cite{mescheder2019occupancy, chen2019learning} or signed distance function \cite{park2019deepsdf, michalkiewicz2019implicit, duan2020curriculum}, which capture more complex geometries. Apart from their strong representation power, previous works show that implicit models encode shape priors in the decoder space, which are able to reconstruct complete shapes from partial observations \cite{park2019deepsdf, duan2020curriculum}. Inspired by this, we propose an implicit function based method to learn to recover clean point clouds from the attacked ones.

\begin{figure*}[tb]
    \centering
    \vspace{-0.2cm}
    %\vspace{-10pt}
    \subfigure[Input Point Cloud]{
        \includegraphics[width=0.17\textwidth]{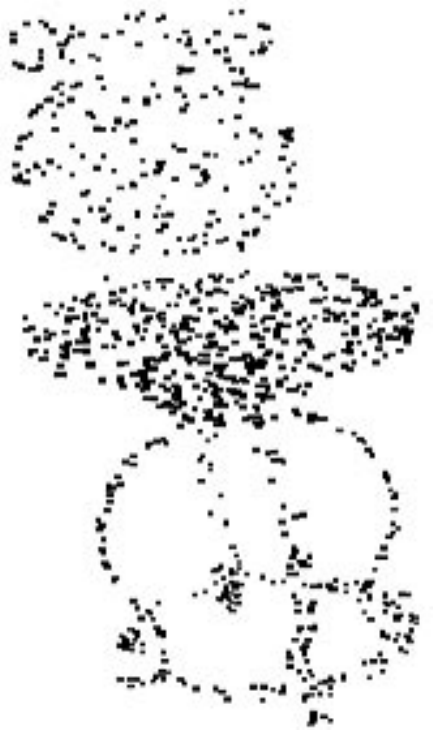}
    }
   \hspace{0.4cm}
    \subfigure[Reconstruction]{
        \includegraphics[width=0.19\textwidth]{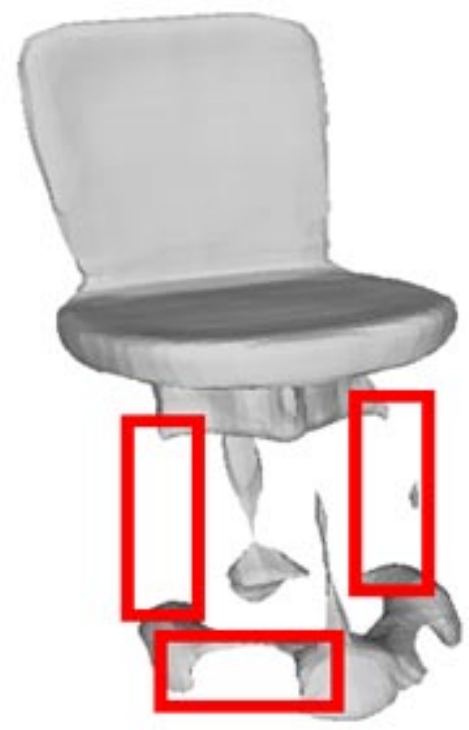}
    }
   \hspace{0.4cm}
    \subfigure[Re-sampled Points]{
        \includegraphics[width=0.18\textwidth]{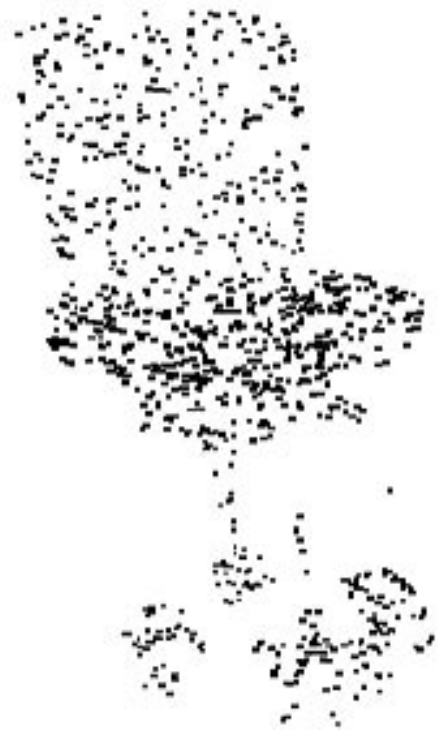}
    }
  \hspace{0.4cm}
    \subfigure[Optimized Points]{
        \includegraphics[width=0.18\textwidth]{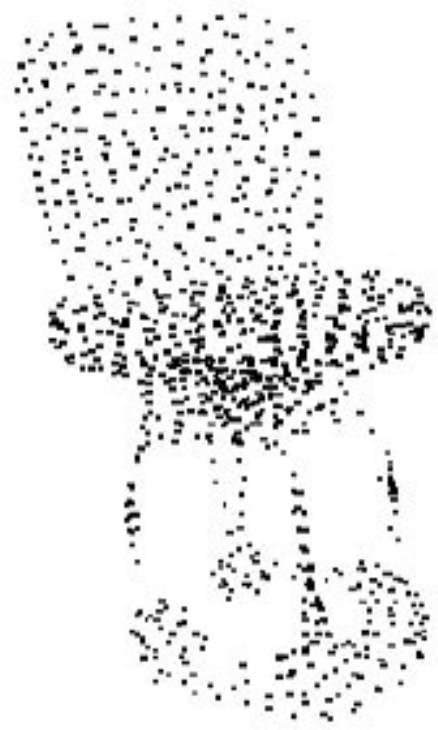}
    }
    \caption{Comparison of the re-meshing and optimization based IF-Defense. Given the (a) input point cloud, the (b) reconstructed mesh using Marching Cubes according to the implicit field fails to capture the chair's legs. As a result, the (c) re-sampled point cloud is misclassified as a monitor by PointNet. In contrast, the (d) optimized point cloud successfully retain the legs and is classified correctly.}
    \label{remesh_vs_deform}
    \vspace{-10pt}
    %\vspace{-0.2cm}
\end{figure*}

\section{Approach}

In this section, we first introduce an intuitive point cloud restoration method by re-meshing and re-sampling. Then, we propose a learning algorithm to directly optimize the coordinates of input points to further improve the robustness of IF-Defense. Finally, we present the implementation details of our method.

\subsection{Re-meshing based IF-Defense}

In order to restore the clean point cloud from existing ones, an intuitive idea is to reconstruct meshes from noisy point cloud at first, and then re-sample points on the mesh. Inspired by the recent success of implicit functions, we employ Occupancy Networks (ONet) \cite{mescheder2019occupancy} and Convolutional Occupancy Networks (ConvONet) \cite{peng2020convolutional} for 3D shape reconstruction. 
The encoder of these networks takes a point cloud as input to obtain a shape latent code, while the decoder outputs the learned implicit fields by querying 3D coordinates. Using the trained implicit function networks, we estimate the implicit surface of the point cloud pre-processed by SOR. 
As the implicit model is pre-trained on clean data, the output space of the decoder lies on the complete and accurate shape manifold, which is beneficial to eliminate the attack of out-of-surface geometric changes. Given the implicit representation of the recovered surface, the next step is to restore the original clean point cloud which reverses the attack effects. We explicitly reconstruct the shape as a mesh using Marching Cubes \cite{lorensen1987marching} algorithm, and then sample from the mesh using the same point sampling method as in training data to get the restored point cloud.

\subsection{Optimization based IF-Defense}

The re-meshing based IF-Defense heavily relies on the quality of the reconstructed meshes, where we sample the restored point clouds. However, even for the recent implicit function based methods, it is still very challenging to reconstruct accurate meshes from the noisy point clouds. Previous studies show that some geometries such as slender parts of an object are difficult to be captured by implicit functions \cite{duan2020curriculum}. Also, the noise in the attacked point cloud may lead to imprecise shape latent codes, which further enlarge the reconstruction errors. For example, ONet \cite{mescheder2019occupancy} fails to reconstruct the legs of a chair in Figure \ref{remesh_vs_deform} (b). As a result, the re-sampled point cloud in Figure \ref{remesh_vs_deform} (c) is misclassified by PointNet as a monitor. This fact motivates us to design a learning based point cloud restoration algorithm that directly optimizes the coordinates of the points rather than a two-step process of re-meshing and re-sampling.

More specifically, we first initialize the defense point cloud $\hat{X}$ as the input. Since the number of the input points may differ from the clean point clouds, we randomly duplicate or sub-sample points in $\hat{X}$ to maintain the same number of points as the training data. 
Then, instead of reconstructing meshes from the implicit field, we directly learn the coordinates of clean point clouds based on the predicted implicit surface by optimizing two losses: geometry-aware loss and distribution-aware loss.

% Tables of results on PN and PN++
\begin{table*}[tb]
    \caption{Classification accuracy of ModelNet40 under various attack and defense methods on PointNet. We report the average Chamfer distance (CD) between clean point clouds and their adversarial counterparts for reference.}
    \label{pn_all_defense}
    \centering
\vspace{5pt}
\renewcommand\arraystretch{1.2}
\begin{tabular}{|c|c|c|c|c|c|c|c|c|c|}
    \hline
    \textbf{Defenses} & \textbf{Clean} & \textbf{Perturb} & \textbf{Add-CD} & \textbf{Add-HD} & \textbf{$k$NN} & \textbf{Drop-100} & \textbf{Drop-200} & \textbf{LG-GAN} & \textbf{AdvPC}\\
    \hline
    CD ($\times10^{-3}$) & - & $0.87$ & $0.88$ & $1.26$ & $1.42$ & $1.65$ & $4.66$ & $8.65$ & $1.10$\\
    \hline
    No defense & $\pmb{88.41\%}$ & $0.00\%$ & $0.00\%$ & $0.00\%$ & $8.51\%$ & $64.67\%$ & $40.24\%$ & $4.40\%$ & $0.00\%$\\
    SRS & $87.44\%$ & $77.47\%$ & $76.34\%$ & $73.66\%$ & $57.41\%$ & $63.57\%$ & $39.51\%$ & $11.72\%$ & $49.01\%$\\
    SOR & $87.88\%$ & $82.81\%$ & $82.58\%$ & $82.25\%$ & $76.63\%$ & $64.75\%$ & $42.59\%$ & $34.90\%$ & $75.45\%$\\
    SOR-AE & $88.09\%$ & $79.86\%$ & $80.15\%$ & $79.58\%$ & $78.28\%$ & $72.53\%$ & $48.06\%$ & $38.56\%$ & $76.60\%$\\
    Adv Training & $88.29\%$ & $25.37\%$ & $19.33\%$ & $15.69\%$ & $19.21\%$ & $70.14\%$ & $49.03\%$ & $4.95\%$ & $12.38\%$\\
    DUP-Net & $87.76\%$ & $84.56\%$ & $83.63\%$ & $82.16\%$ & $80.31\%$ & $67.30\%$ & $46.92\%$ & $35.81\%$ & $77.55\%$\\
    \hline
    \textbf{Ours-Mesh$^\dagger$} & $83.95\%$ & $83.31\%$ & $84.76\%$ & $83.79\%$ & $84.28\%$ & $\pmb{77.76\%}$ & $\pmb{66.94\%}$ & $50.00\%$ & $75.62\%$\\
    \textbf{Ours-Opt$^\dagger$} & $87.07\%$ & $85.78\%$ & $85.94\%$ & $85.94\%$ & $86.18\%$ & $77.63\%$ & $65.28\%$ & $\pmb{52.10\%}$ & $80.14\%$\\
    \textbf{Ours-Opt$^\ddagger$} & $87.64\%$ & $\pmb{86.30\%}$ & $\pmb{86.83\%}$ & $\pmb{86.75\%}$ & $\pmb{86.95\%}$ & $77.39\%$ & $64.63\%$ & $48.11\%$ & $\pmb{80.72\%}$\\
    \hline
\end{tabular}
    \vspace{-5pt}
\end{table*}

\begin{table*}[t]
    \caption{Classification accuracy of ModelNet40 under various attack and defense methods on PointNet++.}
    \label{pn2_all_defense}
    \centering
\vspace{5pt}
\renewcommand\arraystretch{1.2}
\begin{tabular}{|c|c|c|c|c|c|c|c|c|c|}
    \hline
    \textbf{Defenses} & \textbf{Clean} & \textbf{Perturb} & \textbf{Add-CD} & \textbf{Add-HD} & \textbf{$k$NN} & \textbf{Drop-100} & \textbf{Drop-200} & \textbf{LG-GAN} & \textbf{AdvPC}\\
    \hline
    CD ($\times10^{-3}$) & - & $1.14$ & $2.78$ & $3.55$ & $1.93$ & $1.19$ & $2.67$ & $6.45$ & $1.54$\\
    \hline
    No defense & $\pmb{89.34\%}$ & $0.00\%$ & $7.24\%$ & $6.59\%$ & $0.00\%$ & $80.19\%$ & $68.96\%$ & $10.12\%$ & $0.56\%$\\
    SRS & $83.59\%$ & $73.14\%$ & $65.32\%$ & $43.11\%$ & $49.96\%$ & $64.51\%$ & $39.63\%$ & $7.94\%$ & $48.37\%$\\
    SOR & $86.95\%$ & $77.67\%$ & $72.90\%$ & $72.41\%$ & $61.35\%$ & $74.16\%$ & $69.17\%$ & $11.11\%$ & $66.26\%$\\
    SOR-AE & $88.45\%$ & $78.73\%$ & $73.38\%$ & $71.19\%$ & $78.73\%$ & $76.66\%$ & $68.23\%$ & $15.19\%$ & $68.29\%$\\
    Adv Training & $89.10\%$ & $20.03\%$ & $12.27\%$ & $10.06\%$ & $8.63\%$ & $80.39\%$ & $67.14\%$ & $11.25\%$ & $6.49\%$\\
    DUP-Net & $85.78\%$ & $80.63\%$ & $75.81\%$ & $72.45\%$ & $74.88\%$ & $76.38\%$ & $72.00\%$ & $14.76\%$ & $64.76\%$\\
    \hline
    \textbf{Ours-Mesh$^\dagger$} & $83.27\%$ & $81.65\%$ & $77.71\%$ & $79.13\%$ & $72.57\%$ & $82.46\%$ & $72.93\%$ & $18.96\%$ & $65.97\%$\\
    \textbf{Ours-Opt$^\dagger$} & $87.64\%$ & $85.21\%$ & $78.44\%$ & $73.87\%$ & $85.37\%$ & $79.38\%$ & $75.12\%$ & $\pmb{21.38\%}$ & $74.63\%$\\
    \textbf{Ours-Opt$^\ddagger$} & $89.02\%$ & $\pmb{86.99\%}$ & $\pmb{80.19\%}$ & $\pmb{76.09\%}$ & $\pmb{85.62\%}$ & $\pmb{84.56\%}$ & $\pmb{79.09\%}$ & $17.52\%$ & $\pmb{77.06\%}$\\
    \hline
\end{tabular}
    \vspace{-10pt}
\end{table*}

\textbf{Geometry-aware loss} aims to encourage the optimized points to lie on the shape surface. At each time, we concatenate the shape latent code and the coordinate of a point as input to the implicit function, where the output shows the predicted occupancy probability at that point. Then, we employ the binary cross-entropy loss to force the optimized points to approach the surface as follows:
\begin{equation}
    \mathcal{L}_G = \sum_{i=1}^N \mathcal{L}_{ce}(f_\theta(\bm{z}, \bm{x_i}), \tau),
\end{equation}
where $\bm{z}$ is the shape latent code extracted from the input point cloud, $\bm{x_i}$ is the point coordinate to be optimized, and $N$ is the number of points. $f_\theta(\bm{z}, \bm{x_i})$ is the implicit function that outputs the occupancy probability at location $\bm{x_i}$. $\tau$ is a hyper-parameter controlling the object boundary, which is used as the ground-truth occupancy probability of surface. 
For points in $\hat{X}$ that are close to surface, the gradient of geometry-aware loss drives them towards the object boundary. In contrast, for points initialized at the missing parts of the implicit field, the loss provides no gradient since the occupancy probabilities are nearly the same among those regions. Therefore, these points still remain in the missing parts which compensate the errors in the implicit surface.

\textbf{Distribution-aware loss} maximizes the distance from a point to its $k$-nearest neighbors ($k$NN), which encourages a more uniform point distribution:
\begin{equation}
    \mathcal{L}_{D} = \sum_{i=1}^N \sum_{\bm{x_j} \in knn(\bm{x_i}, k)} -||\bm{x_i} - \bm{x_j}|| \cdot e^{-||\bm{x_i} - \bm{x_j}||^2 / h^2},
\end{equation}
where $knn(\bm{x_i}, k)$ denotes the $k$NN of a point $\bm{x_i}$. The exponential term especially punishes the points that are too close to each other, and $h$ is a hyper-parameter controlling the decay rate with respect to the distance. Similar penalization has also been proposed in the previous point up-sampling work \cite{yu2018pu}, known as the repulsion loss. 
We optimize the point coordinates $\bm{x_i}$ by minimizing the following objective function with a hyper-parameter $\lambda$ balancing the weights of two terms:
\begin{equation} \label{optimize}
    \mathcal{L}(\hat{X}) = \mathcal{L}_G + \lambda \mathcal{L}_{D}.
\end{equation}

\subsection{Implementation Details}

We implemented the implicit function network with the widely-used ONet \cite{mescheder2019occupancy} and ConvONet \cite{peng2020convolutional} in IF-Defense, which are trained on multiple categories without class labels. We first pre-trained them on the ShapeNet dataset \cite{chang2015shapenet} and then finetuned them on the ModelNet40 dataset \cite{wu20153d}. For the optimization based IF-Defense, we used $\tau=0.2$ as suggested by \cite{mescheder2019occupancy}. Parameters $h$ and $k$ were set to be $0.03$ and $5$ following \cite{yu2018pu}, and $\lambda$ was set as $500$. We optimized the coordinates of points for 200 iterations using the Adam \cite{kingma2014adam} optimizer with a learning rate equals to $0.01$.

% Table of results on DGCNN and PC
\begin{table*}[tb]
	\vspace{-2pt}
    \caption{Classification accuracy of ModelNet40 under various attack and defense methods on DGCNN.}
    \label{dgcnn_all_defense}
    \centering
\vspace{5pt}
\renewcommand\arraystretch{1.2}
\begin{tabular}{|c|c|c|c|c|c|c|c|c|c|}
    \hline
    \textbf{Defenses} & \textbf{Clean} & \textbf{Perturb} & \textbf{Add-CD} & \textbf{Add-HD} & \textbf{$k$NN} & \textbf{Drop-100} & \textbf{Drop-200} & \textbf{LG-GAN} & \textbf{AdvPC}\\
    \hline
    CD ($\times10^{-3}$) & - & $2.50$ & $3.77$ & $6.97$ & $3.03$ & $1.42$ & $4.36$ & $9.61$ & $2.48$\\
    \hline
    No defense & $\pmb{91.49\%}$ & $0.00\%$ & $1.46\%$ & $1.42\%$ & $20.02\%$ & $75.16\%$ & $55.06\%$ & $15.41\%$ & $9.23\%$\\
    SRS & $81.32\%$ & $50.20\%$ & $63.82\%$ & $43.35\%$ & $41.25\%$ & $49.23\%$ & $23.82\%$ & $20.07\%$ & $41.62\%$\\
    SOR & $88.61\%$ & $76.50\%$ & $72.53\%$ & $63.74\%$ & $55.92\%$ & $64.68\%$ & $59.36\%$ & $30.82\%$ & $56.49\%$\\
    SOR-AE & $89.20\%$ & $79.05\%$ & $76.38\%$ & $66.25\%$ & $56.78\%$ & $66.78\%$ & $63.70\%$ & $32.96\%$ & $58.67\%$\\
    Adv Training & $90.22\%$ & $11.87\%$ & $6.59\%$ & $6.33\%$ & $15.96\%$ & $75.45\%$ & $55.43\%$ & $15.21\%$ & $18.37\%$\\
    DUP-Net & $53.54\%$ & $42.67\%$ & $44.94\%$ & $33.02\%$ & $35.45\%$ & $44.45\%$ & $36.02\%$ & $21.38\%$ & $29.38\%$\\
    \hline
    \textbf{Ours-Mesh$^\dagger$} & $83.91\%$ & $81.56\%$ & $81.73\%$ & $67.50\%$ & $79.38\%$ & $78.97\%$ & $70.34\%$ & $46.09\%$ & $65.54\%$\\
    \textbf{Ours-Opt$^\dagger$} & $88.25\%$ & $82.25\%$ & $81.77\%$ & $67.75\%$ & $82.29\%$ & $79.25\%$ & $\pmb{73.30\%}$ & $\pmb{53.08\%}$ & $76.01\%$\\
    \textbf{Ours-Opt$^\ddagger$} & $89.22\%$ & $\pmb{85.53\%}$ & $\pmb{84.20\%}$ & $\pmb{72.93\%}$ & $\pmb{82.33\%}$ & $\pmb{83.43\%}$ & $73.22\%$ & $50.70\%$ & $\pmb{79.14\%}$\\
    \hline
\end{tabular}
    \vspace{-5pt}
\end{table*}

\begin{table*}[t]
    \caption{Classification accuracy of ModelNet40 under various attack and defense methods on PointConv.}
    \label{pc_all_defense}
    \centering
\vspace{5pt}
\renewcommand\arraystretch{1.2}
\begin{tabular}{|c|c|c|c|c|c|c|c|c|c|}
    \hline
    \textbf{Defenses} & \textbf{Clean} & \textbf{Perturb} & \textbf{Add-CD} & \textbf{Add-HD} & \textbf{$k$NN} & \textbf{Drop-100} & \textbf{Drop-200} & \textbf{LG-GAN} & \textbf{AdvPC}\\
    \hline
    CD ($\times10^{-3}$) & - & $1.14$ & $1.22$ & $1.97$ & $2.27$ & $1.46$ & $4.31$ & $9.66$ & $4.43$\\
    \hline
    No defense & $\pmb{88.49\%}$ & $0.00\%$ & $0.54\%$ & $0.68\%$ & $3.12\%$ & $77.96\%$ & $64.02\%$ & $4.42\%$ & $6.45\%$\\
    SRS & $85.23\%$ & $76.22\%$ & $71.31\%$ & $61.98\%$ & $55.75\%$ & $69.45\%$ & $48.87\%$ & $5.10\%$ & $37.62\%$\\
    SOR & $87.28\%$ & $79.25\%$ & $82.41\%$ & $72.73\%$ & $26.13\%$ & $77.63\%$ & $63.78\%$ & $5.48\%$ & $51.75\%$\\
    SOR-AE & $87.40\%$ & $78.08\%$ & $77.27\%$ & $74.55\%$ & $56.50\%$ & $72.45\%$ & $60.37\%$ & $8.64\%$ & $50.96\%$\\
    Adv Training & $88.90\%$ & $16.57\%$ & $8.32\%$ & $4.84\%$ & $15.64\%$ & $81.00\%$ & $72.33\%$ & $5.25\%$ & $16.20\%$\\
    DUP-Net & $78.73\%$ & $68.84\%$ & $72.61\%$ & $61.14\%$ & $43.76\%$ & $70.75\%$ & $58.23\%$ & $5.02\%$ & $49.35\%$\\
    \hline
    \textbf{Ours-Mesh$^\dagger$} & $82.78\%$ & $81.73\%$ & $81.85\%$ & $75.61\%$ & $77.15\%$ & $75.97\%$ & $68.44\%$ & $15.46\%$ & $53.81\%$\\
    \textbf{Ours-Opt$^\dagger$} & $86.10\%$ & $83.55\%$ & $83.95\%$ & $76.86\%$ & $80.47\%$ & $78.85\%$ & $70.34\%$ & $\pmb{18.78\%}$ & $\pmb{61.77\%}$\\
    \textbf{Ours-Opt$^\ddagger$} & $88.21\%$ & $\pmb{86.67\%}$ & $\pmb{85.62\%}$ & $\pmb{82.13\%}$ & $\pmb{81.08\%}$ & $\pmb{81.20\%}$ & $\pmb{74.51\%}$ & $16.55\%$ & $59.82\%$\\
    \hline
\end{tabular}
    \vspace{-10pt}
\end{table*}

\section{Experiments}

We conducted all the experiments on the commonly used ModelNet40 shape classification benchmark \cite{wu20153d} which contains 12,311 CAD models from 40 man-made object classes. We used the official split with 9,843 shapes for training and 2,468 for testing. Following \cite{qi2017pointnet}, we uniformly sampled $1024$ points from the surface of each object and normalize them into a unit sphere. We applied PointNet \cite{qi2017pointnet}, PointNet++ \cite{qi2017pointnet++}, DGCNN \cite{wang2019dynamic}, PointConv \cite{wu2019pointconv} and RS-CNN \cite{liu2019relation} as the victim models, with the single scale grouping (SSG) strategy for PointNet++ and PointConv.

For the attack methods, we employed the point perturbation and individual point adding attack \cite{xiang2019generating}, $k$NN attack \cite{tsai2020robust}, point dropping attack \cite{zheng2019pointcloud} as well as two recently proposed attacks LG-GAN \cite{zhou2020lg} and AdvPC \cite{hamdi2020advpc}. For the defense baselines, we employed SRS \cite{yang2019adversarial}, SOR \cite{zhou2019dup}, DUP-Net \cite{zhou2019dup} and adversarial training. For adversarial training, all victim models are trained on both clean data and adversarial examples generated by point perturbation. We also trained a point cloud AE \cite{achlioptas2018learning} with a SOR pre-processor as a baseline defense called SOR-AE. We include more details about the implementation of baseline methods in the appendix. Following previous works, we tested on targeted attack and reported the classification accuracy after defense, where higher accuracy indicates better defense.

% ablation study on \lambda
\begin{figure*}[tb]
	\centering
	\vspace{-0.2cm}
	\includegraphics[width=0.9\textwidth]{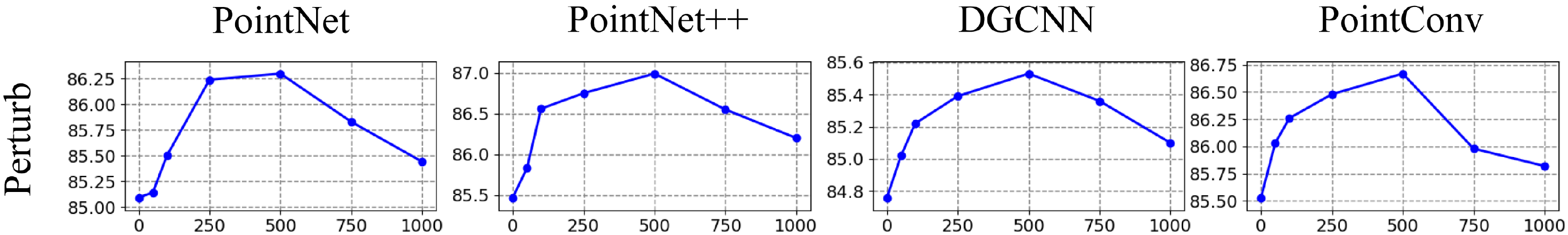}

	\caption{Ablation study of $\lambda$. We show the defense accuracy of four victim models against point perturbation attack, where all four best results are achieved at $\lambda=500$. The defense module evaluated here is the optimization based IF-Defense with ConvONet.}
	\label{ablation_conv_paper}
    \vspace{-6pt}
\end{figure*}

% transfer/black-box attack results
\begin{table*}[tb]
    \caption{Classification accuracy of ModelNet40 under black-box attacks and defenses.}
    \label{transfer}
    \centering
 \vspace{5pt}
\renewcommand\arraystretch{1.2}
\begin{tabular}{|c|c|c|c|c|c|c|}
    \hline
    \textbf{Network} & \textbf{Defense} & \textbf{Add-CD} & \textbf{$k$NN} & \textbf{Drop-200} & \textbf{LG-GAN} & \textbf{AdvPC}\\
    \hline
    \textbf{PointNet} & No defense & $0.00\%$ & $8.51\%$ & $40.24\%$ & $4.40\%$ & $0.00\%$\\
    \hline
    \multirow{4}{*}{\textbf{PointNet++}} & No defense & $87.60\%$ & $80.47\%$ & $79.90\%$ & $24.18\%$ & $70.07\%$\\
    & SOR & $87.13\%$ & $85.07\%$ & $74.84\%$ & $48.78\%$ & $74.09\%$\\
    & DUP-Net & $87.12\%$ & $84.04\%$ & $73.06\%$ & $50.90\%$ & $72.94\%$\\
    & \textbf{Ours-Opt$^\ddagger$} & $\pmb{88.17\%}$ & $\pmb{85.98\%}$ & $\pmb{79.98\%}$ & $\pmb{54.85\%}$ & $\pmb{80.59\%}$\\
    \hline
    \multirow{4}{*}{\textbf{DGCNN}} & No defense & $78.24\%$ & $80.19\%$ & $73.14\%$ & $35.12\%$ & $74.51\%$\\
    & SOR & $85.58\%$ & $87.16\%$ & $66.57\%$ & $40.23\%$ & $78.49\%$\\
    & DUP-Net & $53.20\%$ & $49.47\%$ & $35.01\%$ & $20.35\%$ & $38.77\%$\\
    & \textbf{Ours-Opt$^\ddagger$} & $\pmb{88.09\%}$ & $\pmb{88.01\%}$ & $\pmb{76.90\%}$ & $\pmb{62.13\%}$ & $\pmb{85.61\%}$\\
    \hline
    \multirow{4}{*}{\textbf{PointConv}} & No defense & $84.81\%$ & $77.11\%$ & $76.26\%$ & $22.41\%$ & $64.19\%$\\
    & SOR & $84.57\%$ & $82.43\%$ & $72.41\%$ & $47.52\%$ & $70.89\%$\\
    & DUP-Net & $79.74\%$ & $75.20\%$ & $57.37\%$ & $32.15\%$ & $66.78\%$\\
    & \textbf{Ours-Opt$^\ddagger$} & $\pmb{87.76\%}$ & $\pmb{86.55\%}$ & $\pmb{77.19\%}$ & $\pmb{56.25\%}$ & $\pmb{76.69\%}$\\
    \hline
\end{tabular}
    \vspace{-10pt}
\end{table*}

\subsection{Comparison with the State-of-the-art Methods}

Table \ref{pn_all_defense} and Table \ref{pn2_all_defense} illustrate the classification accuracy under various attack and defense methods on PointNet and PointNet++. In the Tables, Ours-Mesh and Ours-Opt represent the re-meshing and optimization based IF-Defense, respectively. We use $\dagger$ and $\ddagger$ to show the results of two implicit function networks ONet and ConvONet. 
We observe that the optimization based method consistently outperforms the re-meshing based method, showing the effectiveness of our learning based point cloud restoration algorithm. Also, employing ConvONet usually leads to better or comparable accuracy compared with ONet because of the stronger representation capacity of ConvONet. 
Besides, although a point cloud AE can also project the distorted input data to the natural shape manifold \cite{hamdi2020advpc}, it fails to reconstruct point clouds with desired point distribution, and thus performs worse than IF-Defense. 
In addition, though adversarial training improves the robustness against point perturbation, other attack effects can still easily break it\footnote{Note that adversarial training is a white-box defense, while other methods are gray-box because attacker is unaware of the defense module.}.

Overall, IF-Defense achieves relatively small improvements against point perturbation and adding attack compared with existing methods, because these attacks mainly result in out-of-surface perturbation and can be alleviated by SOR. However, our method boosts the performance significantly for $k$NN, point dropping, LG-GAN and AdvPC attack since they mainly introduce on-surface perturbation or significant surface distortion, while IF-Defense can recover natural shape surface via implicit function network and learn to restore point clouds with desired distribution.

As shown in Table \ref{dgcnn_all_defense} and Table \ref{pc_all_defense}, we draw similar observations for DGCNN and PointConv. The optimization based IF-Defense still outperforms its re-meshing based counterpart, and ConvONet demonstrates competitive or superior performance compared with ONet. 
It is worth noticing that DUP-Net performs poorly on these two models. This is because DGCNN and PointConv are sensitive to local point distributions as they propagate features through $k$NN graphs. However, DUP-Net up-samples points to a much higher density, which largely affects the learned local graphs due to the difference in point distributions. Instead, the proposed IF-Defense learns uniform point distribution, which leads to better $k$NN graphs. Therefore, we achieve significantly better results than DUP-Net against all the attacks on DGCNN and PointConv. Due to the limited space, we leave the defense results on RS-CNN to the appendix, where we have a similar observation.

\subsection{Ablation Study}

\textbf{Distribution-aware loss weight.} In this part, we study the effect of the hyper-parameter $\lambda$ of the optimization based IF-Defense (Ours-Opt), where ConvONet is adopted as it achieves the best performance against most of the attacks. We varied $\lambda$ between $0$ and $1,000$ and recorded the accuracy of the victim models after defense. As shown in Figure \ref{ablation_conv_paper}, with the increase of $\lambda$, the accuracy first improves and then begins to decrease. In most cases, we observe that the best accuracy is achieved at $\lambda=500$. The distribution-aware loss enforces the points to distribute uniformly over the surface. The points are not able to cover the entire object uniformly with a small $\lambda$, while a large $\lambda$ fails to capture the surface precisely due to the ignorance of the geometry-aware loss. To this end, we select a proper $\lambda$ to balance the importance between accurate surfaces and uniform point distributions. More ablation studies on $\lambda$ using ONet or against other attacks are provided in the appendix.

% results of adaptive attack
\begin{table}[tb]
    \footnotesize
    \caption{Classification accuracy of ModelNet40 under adaptive attack against IF-Defense.}
    \label{adaptive_attack}
    \centering
 \vspace{5pt}
\renewcommand\arraystretch{1.2}
\begin{tabular}{|c|c|c|c|c|}
    \hline
    \textbf{Defenses} & \textbf{PointNet} & \textbf{PointNet++} & \textbf{DGCNN} & \textbf{PointConv}\\
    \hline
    \textbf{Ours-Mesh$^\dagger$} & $56.60\%$ & $52.71\%$ & $55.67\%$ & $52.27\%$\\
    \textbf{Ours-Opt$^\dagger$} & $60.53\%$ & $56.20\%$ & $58.67\%$ & $55.67\%$\\
    \textbf{Ours-Opt$^\ddagger$} & $\pmb{65.90\%}$ & $\pmb{63.80\%}$ & $\pmb{61.81\%}$ & $\pmb{60.07\%}$\\
    \hline
\end{tabular}
    \vspace{-15pt}
\end{table}

% visualization
\begin{figure*}[t]
	\centering
	\vspace{-10pt}
	\includegraphics[width=0.98\textwidth]{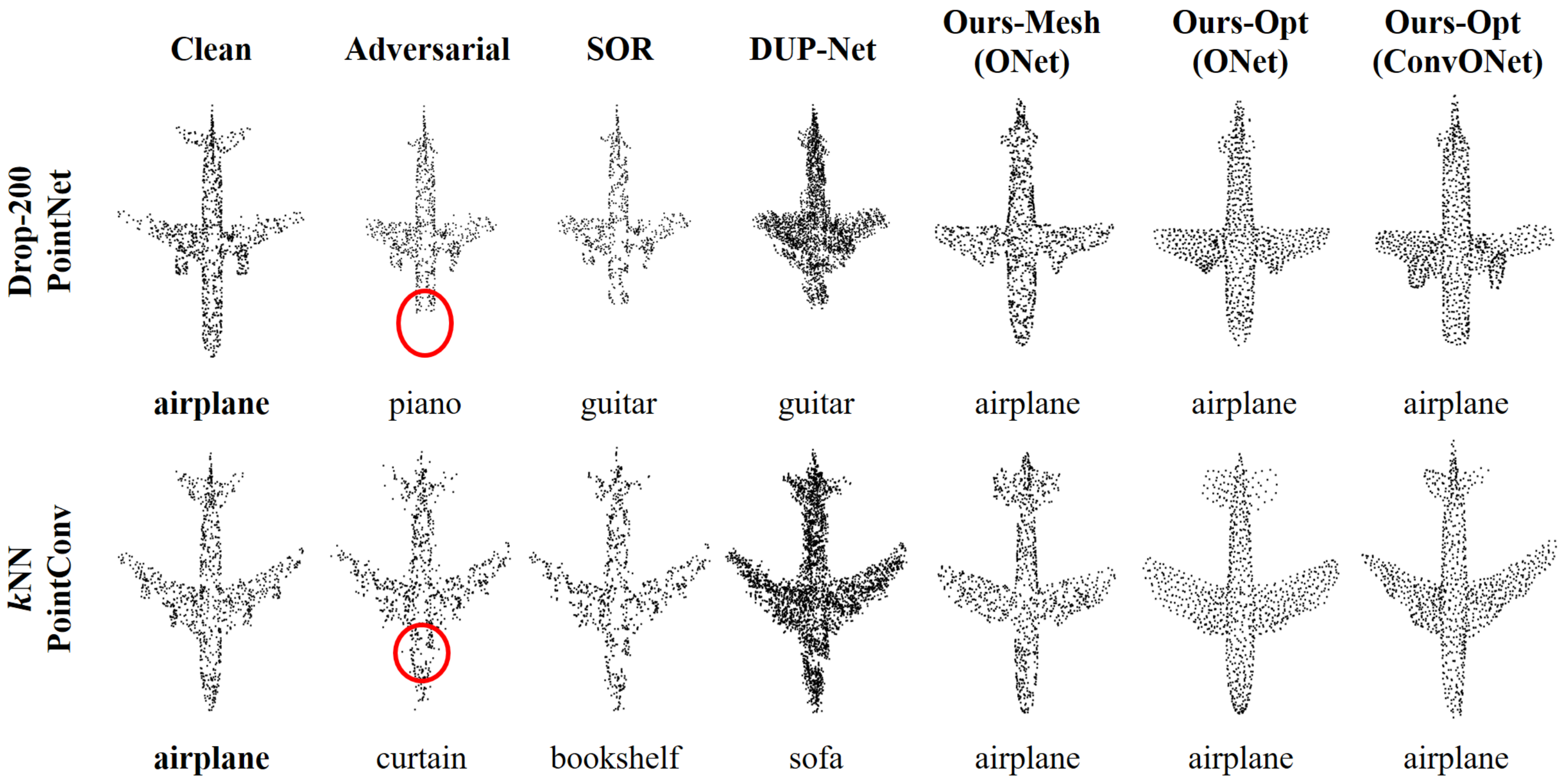}

	\caption{Visualization of different defense results. The labels under each point cloud are the prediction outputs of the victim models.}
	\label{vis_defense_main}
    \vspace{-10pt}
\end{figure*}

\textbf{Adaptive attack.} As pointed out by \cite{carlini2019evaluating}, it is not sufficient to evaluate a proposed defense only against existing attacks. Therefore, we designed an adaptive attack targeting on IF-Defense to study its worst-case bound. Since the SOR pre-processor and the optimization process are non-differentiable, we cannot perform gradient ascent using the cross-entropy loss over the output of the classifier. Instead, we utilize two loss terms to distort the implicit field predicted by the implicit function network. More details about our design are provided in the appendix.

Table \ref{adaptive_attack} summarizes the defense results against the adaptive attack. We controlled the Chamfer distance between clean and adversarial point clouds to be around $2.00$. Comparing to existing attacks, the adaptive attack causes greater performance drop. Nonetheless, Ours-Opt with ConvONet still achieves classification accuracy higher than $60\%$ on all the four victim models. Though the adaptive attack perturbs the shape latent code extracted by the encoder of the implicit function network, the decoder can still project it back to a clean shape, mitigating the distortion.

\subsection{Black-Box Attacks and Defenses}

We explore the transferability of attacks and the performance of various defense methods in this black-box setting. Following previous works \cite{zhou2019dup, zhou2020lg}, we first generated adversarial examples against PointNet, and then transferred them to the other three victim models. We adopted the optimization based IF-Defense with ConvONet for comparison. The results are summarized in Table \ref{transfer}. As the attacked point clouds are generated against PointNet, they are less effective for other network architectures due to the limited transferability. We observe that our method consistently outperforms other defense methods. For SOR and DUP-Net, the classification accuracy even drops in some situations compared with directly using the attacked point clouds. Instead, our IF-Defense continuously boosts the performance, which demonstrates its effectiveness and robustness.

\subsection{Qualitative Results}

Figure \ref{vis_defense_main} illustrates two groups of defense results using SOR, DUP-Net and all three variants of IF-Defense. The first row shows the results under point dropping attack on PointNet, where the head of the airplane is discarded in the adversarial example. SOR fails to defend this adversary because it just removes more points from the point cloud. Although DUP-Net further up-samples the point cloud with PU-Net, the up-sampled points are all near the input points so that the missing part cannot be recovered. Instead, all three IF-Defense methods successfully restore the shape by extending the front end trying to form a head, which demonstrates its effectiveness in reconstructing the whole shapes under partial observations. The second row is the $k$NN attack on PointConv. Most of the points are perturbed along the surface because of the $k$NN constraint, resulting in significant changes in point distribution. DUP-Net fails to recover the original point distribution as it outputs a much denser point cloud. Ours-Mesh re-samples points from the reconstructed mesh using the same sampling strategy as clean data, and Ours-Opt outputs uniformly distributed points because of the distribution-aware loss. Consequently, PointConv correctly classifies the airplane in both cases.

\section{Conclusion}

In this paper, we have proposed a general framework called IF-Defense for adversarial defense in 3D point cloud, which simultaneously addresses point perturbation and surface distortion effects. 
Our IF-Defense learns to restore the clean point clouds by optimizing the coordinates of the attacked points according to geometry-aware and distribution-aware losses, so that the distortion on surfaces is recovered through implicit function and the perturbation on point distributions is eliminated via optimization. 
Extensive experiments show that IF-Defense consistently outperforms existing adversarial defense methods against various 3D point cloud adversarial attacks on PointNet, PointNet++, DGCNN, PointConv and RS-CNN.

\section*{Acknowledgements}

This research was supported by a Vannevar Bush Faculty Fellowship, a grant from the Stanford-Ford Alliance, NSF grant IIS-1763268, and gifts from Adobe, Amazon AWS, and Snap, Inc.

{\small
\bibliographystyle{ieee_fullname}
\bibliography{egbib}
}

\clearpage

\appendix

\section{More Implementation Details about IF-Defense}

\subsection{Training of Implicit Function Networks}

We trained the implicit function networks in the reconstruction task, and then employed them in our IF-Defense. Occupancy Network (ONet) \cite{mescheder2019occupancy} and Convolutional Occupancy Network (ConvONet) \cite{peng2020convolutional} are adopted in our experiments. Specifically, we use the multi-plane encoder and decoder variant of ConvONet (denoted as 2D $3 \times 64^2$ in their paper) because it achieves the best trade-off between reconstruction accuracy and inference speed. We modified their online official code\footnote{https://github.com/autonomousvision/occupancy\_networks} \footnote{https://github.com/autonomousvision/convolutional\_occupancy\_networks} to accommodate our experimental settings. All network structures remain the same as their original papers. In this section, we describe the datasets and training schedules used in our implementations.

\textbf{Datasets.} We used two datasets to train the implicit function networks. For the ShapeNet dataset \cite{chang2015shapenet}, we directly used the processed data provided by \cite{mescheder2019occupancy}, which contains point clouds sampled from each 3D object and their corresponding ground-truth occupancy values. For the ModelNet40 dataset \cite{wu20153d}, we utilized the pre-processing pipeline provided by \cite{mescheder2019occupancy} to generate training data. We also subdivided the training set of ModelNet40 into a training and a validation subset on which we tracked the loss of the models to determine when to stop training as in \cite{mescheder2019occupancy}. It is worth noticing that, we only trained the implicit function networks on the training set of ShapeNet and ModelNet40, so neither the clean data in the test set nor the adversarial point clouds generated by the attacks are leveraged during training. Experimental results on cross dataset evaluation in Section \ref{more_ablation} prove that our IF-Defense is able to generalize to unseen data.

\textbf{Training schedules.} We adopted the same training schedules as their original implementations (e.g. the Adam \cite{kingma2014adam} optimizer with a constant learning rate equals to $1e-4$) except for two aspects. The first one is that we performed random rotation along the z-axis as data augmentation because the object orientations are not aligned in the ModelNet40 dataset. The second aspect is that we set the number of input points as $600$ in ConvONet \cite{peng2020convolutional} compared with the original $3000$ since the adversarial point clouds often consist of only $1024$ points. The training of ONet took for around $2500$k iterations on ShapeNet and $350$k iterations on ModelNet40. For ConvONet, we trained for around $1000$k and $200$k iterations on the two datasets respectively.

\subsection{Hybrid Training of Victim Models}

We applied PointNet \cite{qi2017pointnet}, PointNet++ \cite{qi2017pointnet++}, DGCNN \cite{wang2019dynamic}, PointConv \cite{wu2019pointconv} and RS-CNN \cite{liu2019relation} as the victim models in our experiments. However, we discovered that the victim models trained purely on clean data degraded the accuracy by up to $5\%$ when tested on clean point clouds with the IF-Defense module. In order to eliminate the domain gap between clean point clouds and their defended counterparts, we trained our victim models on both clean and defense point clouds (dubbed hybrid training) and used them to evaluate the performance of IF-Defense.

We argue that this setting is valid in our task. On one hand, the attacks are still conducted against the hybrid trained models under the white-box setting. The post-attack model accuracy only differs less than $2\%$ for each attack and the average Chamfer distance between clean and adversarial point clouds are very close, indicating that the budgets are similar for those attacks. On the other hand, the hybrid training can be regarded as a simple data augmentation. It is completely different from adversarial training \cite{liu2019extending} because the victim models never utilize the adversarial examples during hybrid training.

% defense results for RS-CNN
\begin{table*}[tb]
    \caption{Classification accuracy of ModelNet40 under various attack and defense methods on RS-CNN. We report the average Chamfer distance (CD) between clean point clouds and their adversarial counterparts for reference.}
    \label{rscnn_all_defense}
    \centering
\vspace{5pt}
\renewcommand\arraystretch{1.2}
\begin{tabular}{|c|c|c|c|c|c|c|c|}
    \hline
    \textbf{Defenses} & \textbf{Clean} & \textbf{Perturb} & \textbf{Add-CD} & \textbf{Add-HD} & \textbf{$k$NN} & \textbf{Drop-100} & \textbf{Drop-200}\\
    \hline
    CD ($\times10^{-3}$) & - & $3.49$ & $3.76$ & $4.37$ & $2.67$ & $1.47$ & $4.25$\\
    \hline
    No defense & $\pmb{92.02\%}$ & $0.23\%$ & $4.23\%$ & $3.76\%$ & $2.15\%$ & $73.30\%$ & $56.97\%$\\
    SRS & $91.15\%$ & $75.57\%$ & $62.24\%$ & $48.95\%$ & $56.20\%$ & $72.69\%$ & $53.93\%$\\
    SOR & $90.98\%$ & $82.82\%$ & $82.66\%$ & $79.66\%$ & $70.06\%$ & $75.04\%$ & $64.18\%$\\
    SOR-AE & $90.45\%$ & $80.13\%$ & $78.58\%$ & $80.02\%$ & $76.85\%$ & $77.69\%$ & $70.23\%$\\
    Adv Training & $91.04\%$ & $25.79\%$ & $12.08\%$ & $10.16\%$ & $15.65\%$ & $72.53\%$ & $53.00\%$\\
    DUP-Net & $90.23\%$ & $82.82\%$ & $81.77\%$ & $76.86\%$ & $79.50\%$ & $78.16\%$ & $67.46\%$\\
    \hline
    \textbf{Ours-Mesh$^\dagger$} & $85.29\%$ & $83.52\%$ & $81.38\%$ & $78.74\%$ & $82.98\%$ & $79.46\%$ & $71.60\%$\\
    \textbf{Ours-Opt$^\dagger$} & $89.22\%$ & $86.55\%$ & $84.40\%$ & $\pmb{83.39\%}$ & $84.64\%$ & $80.55\%$ & $71.88\%$\\
    \textbf{Ours-Opt$^\ddagger$} & $91.26\%$ & $\pmb{86.59\%}$ & $\pmb{84.60\%}$ & $81.48\%$ & $\pmb{85.29\%}$ & $\pmb{81.93\%}$ & $\pmb{75.65\%}$\\
    \hline
\end{tabular}
    \vspace{-5pt}
\end{table*}

\section{Baselines}

In order to conduct fair comparisons between IF-Defense and baseline methods, we re-implemented all the attacks \cite{xiang2019generating, zheng2019pointcloud, tsai2020robust} and defenses \cite{yang2019adversarial, zhou2019dup} in PyTorch \cite{paszke2019pytorch} (except for LG-GAN \cite{zhou2020lg} and AdvPC \cite{hamdi2020advpc} that we modified their official implementations to fit in with our experimental settings). Here we detail some of the hyper-parameters and settings in our experiments.

\subsection{Attacks}

Except for point dropping attack \cite{zheng2019pointcloud} that cannot targetedly attack the victim models, we conducted targeted attack for all the other methods. We randomly assigned a target label unequal to the ground-truth class for every example in the ModelNet40 \cite{wu20153d} test set, and this target label assignment was kept unchanged in all the attacks to eliminate the effect of randomness.

\textbf{Perturb.} We followed \cite{xiang2019generating} and used the C\&W attack framework with per-point $L_2$ norm as the perturbation metric. We performed 10-step binary search with 500 iterations in each step.

\textbf{Add.} We followed \cite{xiang2019generating} and used the C\&W attack framework with Chamfer or Hausdorff measurements as the perturbation metrics (dubbed Add-CD and Add-HD respectively). 512 points were added and perturbed during optimization while the original points in the point cloud remain unchanged. We performed 10-step binary search with 500 iterations in each step.

\textbf{$k$NN.} We followed \cite{tsai2020robust} and used the C\&W attack framework with Chamfer measurement and $k$-nearest neighbors ($k$NN) distance as the perturbation metrics. We applied the same clipping and projection operations as their original paper. The optimization iterations were 2,500 in the attack.

\textbf{Drop.} We dropped 100 or 200 points in this attack (dubbed Drop-100 and Drop-200 respectively). Following \cite{zheng2019pointcloud}, we used a greedy algorithm that iteratively calculated the saliency map of all the remaining points and discarded 5 points with the highest saliency scores.

\textbf{LG-GAN.} We modified the online official implementation\footnote{https://github.com/RyanHangZhou/LG-GAN} to perform targeted attack according to our pre-assigned target labels. The hyper-parameters were the same as \cite{zhou2020lg} for all the victim models.

\textbf{AdvPC.} We modified the online official implementation\footnote{https://github.com/ajhamdi/AdvPC} to perform targeted attack according to our pre-assigned target labels. Following \cite{hamdi2020advpc}, 2 different initializations for the optimization were used and the number of iterations was 200 in each optimization. We used the default hyper-parameters provided in their official implementation.

% defense results against randomly dropping points
\begin{table*}[tb]
    \caption{Classification accuracy of sparse point clouds under different defense methods. 600 points are randomly dropped from the original $1024$ points. We list the model accuracy on clean point clouds in the first row for reference. Because input point clouds are sparse, we change the number of dropped points for SRS to 200, and the number of input points for ConvONet to 400.}
    \label{rand_600_defense}
    \centering
\vspace{5pt}
\renewcommand\arraystretch{1.2}
\begin{tabular}{|c|c|c|c|c|c|}
    \hline
    \textbf{Defenses} & \textbf{PointNet} & \textbf{PointNet++} & \textbf{DGCNN} & \textbf{PointConv} & \textbf{RS-CNN}\\
    \hline
    Clean & $88.41\%$ & $89.34\%$ & $91.49\%$ & $88.49\%$ & $92.02\%$\\
    \hline
    No defense & $81.79\%$ & $73.26\%$ & $62.36\%$ & $79.34\%$ & $82.41\%$\\
    SRS & $78.93\%$ & $29.25\%$ & $16.53\%$ & $30.51\%$ & $55.88\%$\\
    SOR & $80.79\%$ & $48.74\%$ & $44.49\%$ & $65.19\%$ & $69.04\%$\\
    SOR-AE & $82.41\%$ & $65.89\%$ & $53.78\%$ & $66.43\%$ & $72.43\%$\\
    Adv Training & $\pmb{86.99\%}$ & $81.52\%$ & $62.24\%$ & $81.16\%$ & $84.97\%$\\
    DUP-Net & $84.16\%$ & $70.46\%$ & $23.22\%$ & $35.78\%$ & $79.25\%$\\
    \hline
    \textbf{Ours-Mesh$^\dagger$} & $82.74\%$ & $82.54\%$ & $83.06\%$ & $80.83\%$ & $84.04\%$\\
    \textbf{Ours-Opt$^\dagger$} & $84.56\%$ & $84.52\%$ & $84.16\%$ & $83.31\%$ & $85.90\%$\\
    \textbf{Ours-Opt$^\ddagger$} & $85.37\%$ & $\pmb{86.43\%}$ & $\pmb{86.14\%}$ & $\pmb{86.14\%}$ & $\pmb{86.43\%}$\\
    \hline
\end{tabular}
    \vspace{-5pt}
\end{table*}

\begin{table*}[tb]
    \caption{Classification accuracy of sparse point clouds under different defense methods. 800 points are randomly dropped from the original $1024$ points. We list the model accuracy on clean point clouds in the first row for reference. Because input point clouds are sparse, we change the number of dropped points for SRS to 100, and the number of input points for ONet and ConvONet to 200.}
    \label{rand_800_defense}
    \centering
\vspace{5pt}
\renewcommand\arraystretch{1.2}
\begin{tabular}{|c|c|c|c|c|c|}
    \hline
    \textbf{Defenses} & \textbf{PointNet} & \textbf{PointNet++} & \textbf{DGCNN} & \textbf{PointConv} & \textbf{RS-CNN}\\
    \hline
    Clean & $88.41\%$ & $89.34\%$ & $91.49\%$ & $88.49\%$ & $92.02\%$\\
    \hline
    No defense & $79.38\%$ & $29.94\%$ & $16.57\%$ & $31.24\%$ & $55.31\%$\\
    SRS & $64.99\%$ & $15.15\%$ & $7.62\%$ & $9.04\%$ & $25.97\%$\\
    SOR & $68.11\%$ & $23.01\%$ & $12.97\%$ & $16.86\%$ & $34.81\%$\\
    SOR-AE & $71.54\%$ & $29.23\%$ & $20.59\%$ & $28.73\%$ & $52.73\%$\\
    Adv Training & $81.48\%$ & $34.97\%$ & $19.00\%$ & $56.28\%$ & $59.04\%$\\
    DUP-Net & $75.32\%$ & $31.89\%$ & $10.41\%$ & $8.27\%$ & $40.92\%$\\
    \hline
    \textbf{Ours-Mesh$^\dagger$} & $81.12\%$ & $79.70\%$ & $81.34\%$ & $79.01\%$ & $81.12\%$\\
    \textbf{Ours-Opt$^\dagger$} & $82.13\%$ & $\pmb{81.44\%}$ & $\pmb{81.36\%}$ & $81.04\%$ & $\pmb{83.35\%}$\\
    \textbf{Ours-Opt$^\ddagger$} & $\pmb{82.37\%}$ & $80.96\%$ & $81.12\%$ & $\pmb{81.40\%}$ & $82.13\%$\\
    \hline
\end{tabular}
    \vspace{-5pt}
\end{table*}

\textbf{Adaptive attack.} In the adaptive attack, we assume the attacker has full knowledge of IF-Defense module. We applied the C\&W attack framework with Chamfer measurement and $k$NN distance as the perturbation metrics because they can introduce robust adversarial examples as shown in $k$NN attack. Since IF-Defense is non-differentiable, we cannot directly attack the victim models' classification output. Instead, we proposed to destroy the predicted implicit field using two loss terms. Let $X=\{\bm{x_i}\}_{i=1}^N$ be the clean point cloud, and $\hat{X}=\{\bm{\hat{x_i}}\}_{i=1}^N$ be its adversarial counterpart initialized by $X$. Let $\bm{z}$ and $\bm{\hat{z}}$ be the shape latent code extracted from $X$ and $\hat{X}$ respectively. The first loss aims at distorting the shape of the point cloud by:
\begin{equation}
    \mathcal{L}_1 = -||\bm{z} - \bm{\hat{z}}||_2^2
\end{equation}
The second loss aims to drive the perturbed points away from the clean object surface:
\begin{equation}
    \mathcal{L}_2 = -\sum_{i=1}^N \mathcal{L}_{ce}(f_\theta(\bm{z}, \bm{\hat{x_i}}), \tau),
\end{equation}
where $f_\theta(\bm{z}, \bm{\hat{x_i}})$ is the implicit function that outputs the occupancy probability at location $\bm{\hat{x_i}}$. $\tau$ is a hyper-parameter controlling the object boundary, which is used as the ground-truth occupancy probability of surface. We set $\tau=0.2$ as in IF-Defense. The complete adversarial loss function is balanced by a hyper-parameter $\alpha$:
\begin{equation}
    \mathcal{L}_{adv} = \mathcal{L}_1 + \alpha \mathcal{L}_2,
\end{equation}
where we used $\alpha=10$ and $\alpha=1$ for IF-Defense with ONet and ConvONet, respectively. The optimization iterations were 2,500 as in \cite{tsai2020robust}.

% ablation on \lambda
\begin{figure*}[tb]
	\centering
	\includegraphics[width=0.9\textwidth]{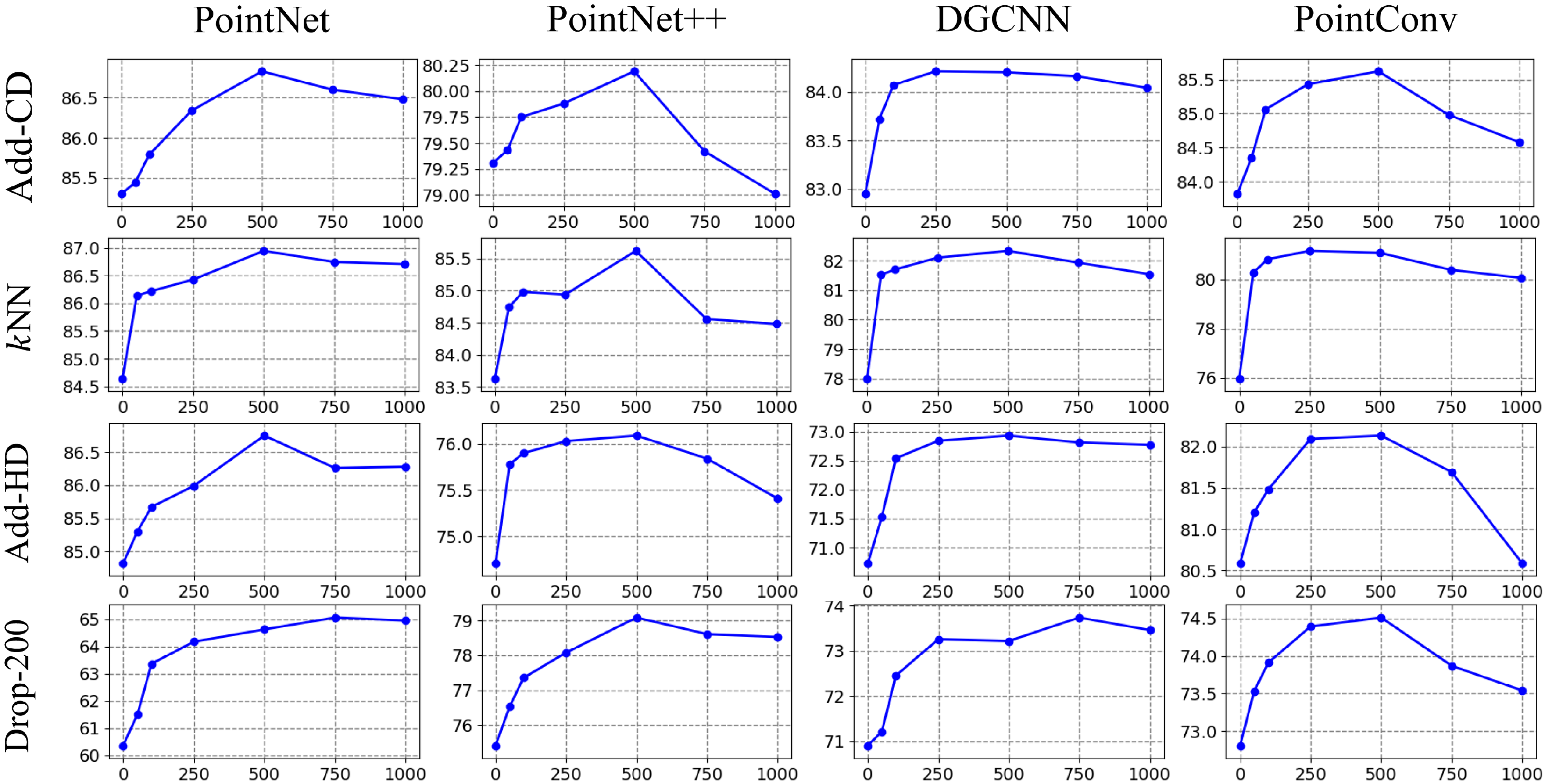}

	\caption{Ablation study of $\lambda$, where we show the defense accuracy of four victim models against point adding attack with Chamfer or Hausdorff measurement, $k$NN attack and point dropping attack. The defense evaluated here is the optimization based IF-Defense with ConvONet.}
	\label{ablation_conv_appendix}
	\vspace{-5pt}
\end{figure*}

\begin{figure*}[tb]
	\centering
	\includegraphics[width=0.9\textwidth]{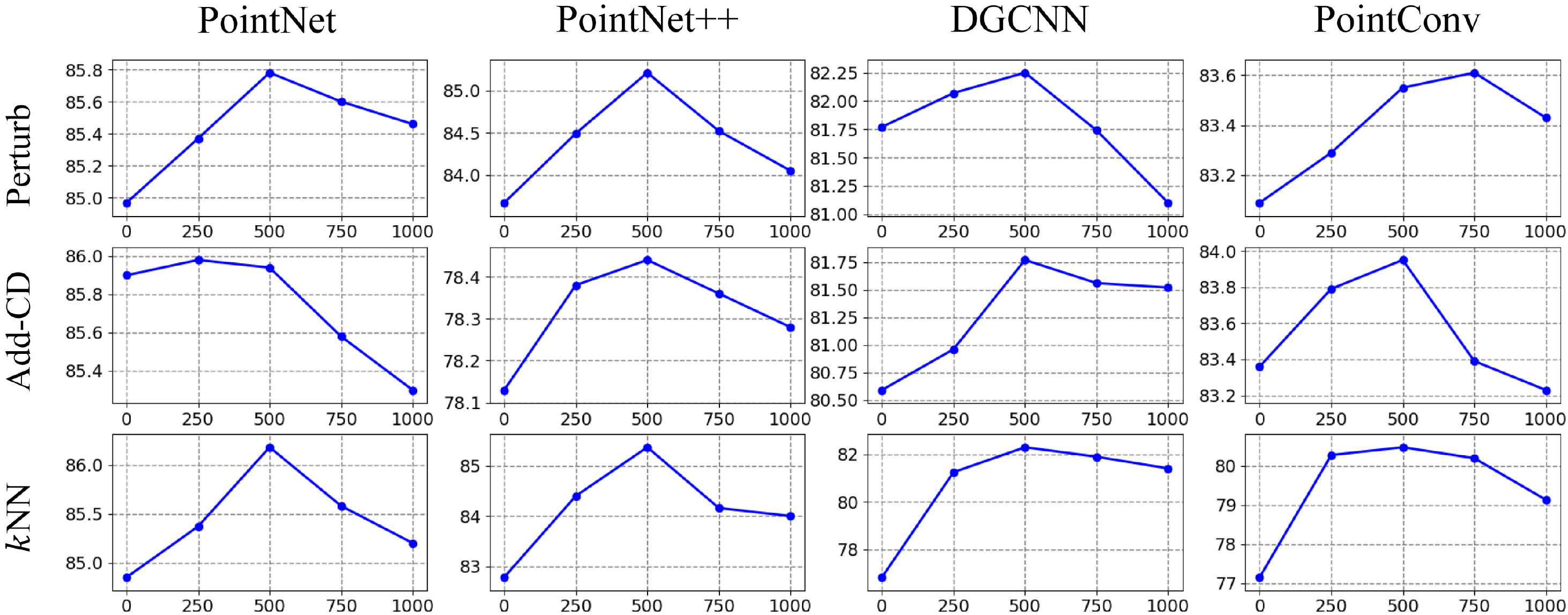}

	\caption{Ablation study of $\lambda$, where we show the defense accuracy of four victim models against point perturbation attack, point adding attack with Chamfer measurement and $k$NN attack. The defense evaluated here is the optimization based IF-Defense with ONet.}
	\label{ablation_onet_appendix}
    \vspace{-8pt}
\end{figure*}

\subsection{Defenses}

\textbf{SRS.} We randomly dropped $500$ points from the input point cloud as done by \cite{zhou2019dup}.

\textbf{SOR.} We trimmed the points in a point cloud $X$ according to $X'=\{x_i|d_i \leq \mu_d + \alpha \cdot \sigma_d\}$, where $\mu_d$ and $\sigma_d$ are the mean and standard deviation of the $k$NN distance of all points in $X$. We used $k=2$ and $\alpha=1.1$ as \cite{zhou2019dup}.

\textbf{DUP-Net.} For the SOR pre-processor, we used $k=2$ and $\alpha=1.1$ as in the individual SOR defense. For the PU-Net \cite{yu2018pu}, we trained it on the Visionair dataset \cite{yu2018pu} using Chamfer distance loss and repulsion loss as \cite{zhou2019dup}. The up-sampling rate was 4 and we used the same training schedules. We also tried first pre-training PU-Net on ShapeNet and then finetuning it on ModelNet40 but observed no performance gain. We conjecture that this is because of the limited capacity of PU-Net. PU-Net is still not able to recover uniform point distributions or natural shapes even with more training data.

\textbf{SOR-AE.} \cite{hamdi2020advpc} leveraged a point cloud auto-encoder (AE) \cite{achlioptas2018learning} as a baseline defense to evaluate the effectiveness of their attack. They showed that this defense can even outperform DUP-Net against some attacks. Therefore, we also adopted point cloud AE as a baseline in our experiments. Additionally, we discovered that adding a SOR pre-processor can further improve its robustness, which is the SOR-AE defense we utilized. We still set $k$ and $\alpha$ as $2$ and $1.1$ in the SOR pre-processor. The point cloud AE is first pre-trained on the ShapeNet dataset and then fine-tuned on the ModelNet40 dataset in the reconstruction task as \cite{hamdi2020advpc}.

\textbf{Adv Training.} We follow \cite{liu2019extending} to train the point cloud networks on both clean and adversarial examples. While it is computationally infeasible to generate adversarial data in all types of attack effects, we leveraged PGD \cite{madry2017towards} attack to craft point perturbation as \cite{liu2019extending}.

\textbf{Other baselines.} We have considered using point cloud denoising networks \cite{rakotosaona2020pointcleannet, pistilli2020learning, hermosilla2019total} as baselines in the adversarial defense task. However, these methods work on local patches, thus requiring the input to be dense point clouds (usually more than $10$k points). This is incompatible with our setting where the number of input points is often $1024$. In addition, we also tried 3D reconstruction networks such as DMC \cite{liao2018dmc} and Point2Mesh \cite{Hanocka2020p2m} for defense. However, they either presents much worse reconstruction results than the implicit function networks, or requires face normals as input. Besides, their explicit output representations prevent their adaptation into the learning based restoration algorithm. Therefore, we do not include them as baselines.

% SOR ablation
\begin{table*}[tb]
    \caption{Ablation study of the SOR pre-processor in IF-Defense. We show the model classification accuracy with or without SOR for comparison. The defense evaluated here is the optimization based IF-Defense with ConvONet.}
    \label{sor_ablation}
    \centering
 \vspace{5pt}
\renewcommand\arraystretch{1.2}
\begin{tabular}{|c|c|c|c|c|c|c|c|c|c|c|}
    \hline
    \textbf{Attack} & \multicolumn{2}{c|}{\textbf{PointNet}} & \multicolumn{2}{c|}{\textbf{PointNet++}} & \multicolumn{2}{c|}{\textbf{DGCNN}} & \multicolumn{2}{c|}{\textbf{PointConv}} & \multicolumn{2}{c|}{\textbf{RS-CNN}}\\
    \hline
    \textbf{SOR} & $\checkmark$ & $\times$ & $\checkmark$ & $\times$ & $\checkmark$ & $\times$ & $\checkmark$ & $\times$ & $\checkmark$ & $\times$\\
    \hline
    \textbf{Perturb} & $\pmb{86.30\%}$ & $83.29\%$ & $\pmb{86.99\%}$ & $83.98\%$ & $\pmb{85.53\%}$ & $82.13\%$ & $\pmb{86.67\%}$ & $83.64\%$ & $\pmb{86.59\%}$ & $84.72\%$\\
    \textbf{Add-CD} & $\pmb{86.83\%}$ & $63.12\%$ & $\pmb{80.19\%}$ & $69.75\%$ & $\pmb{84.20\%}$ & $48.58\%$ & $\pmb{85.62\%}$ & $73.29\%$ & $\pmb{84.60\%}$ & $68.63\%$\\
    \textbf{$k$NN} & $\pmb{86.95\%}$ & $84.44\%$ & $\pmb{85.62\%}$ & $83.18\%$ & $\pmb{82.33\%}$ & $81.44\%$ & $\pmb{81.08\%}$ & $79.74\%$ & $\pmb{85.29\%}$ & $84.12\%$\\
    \textbf{Drop-200} & $64.63\%$ & $\pmb{65.52\%}$ & $79.09\%$ & $\pmb{79.46\%}$ & $73.22\%$ & $\pmb{74.35\%}$ & $74.51\%$ & $\pmb{76.05\%}$ & $75.65\%$ & $\pmb{75.88\%}$\\
    \hline
\end{tabular}
    \vspace{-5pt}
\end{table*}

% cross dataset
\begin{table*}[tb]
    \caption{Cross dataset evaluation for IF-Defense. We show the defense results using ConvONet trained with or without ModelNet40 data for comparison. The defense evaluated here is the optimization based IF-Defense with ConvONet.}
    \label{cross_dataset}
    \centering
 \vspace{5pt}
\renewcommand\arraystretch{1.2}
\begin{tabular}{|c|c|c|c|c|c|c|c|c|c|c|}
    \hline
    \textbf{Attack} & \multicolumn{2}{c|}{\textbf{PointNet}} & \multicolumn{2}{c|}{\textbf{PointNet++}} & \multicolumn{2}{c|}{\textbf{DGCNN}} & \multicolumn{2}{c|}{\textbf{PointConv}} & \multicolumn{2}{c|}{\textbf{RS-CNN}}\\
    \hline
    \textbf{ModelNet40} & $\checkmark$ & $\times$ & $\checkmark$ & $\times$ & $\checkmark$ & $\times$ & $\checkmark$ & $\times$ & $\checkmark$ & $\times$\\
    \hline
    \textbf{Perturb} & $86.30\%$ & $\pmb{86.67\%}$ & $\pmb{86.99\%}$ & $85.98\%$ & $\pmb{85.53\%}$ & $85.13\%$ & $\pmb{86.67\%}$ & $84.72\%$ & $\pmb{86.59\%}$ & $85.94\%$\\
    \textbf{Add-CD} & $86.83\%$ & $\pmb{87.36\%}$ & $\pmb{80.19\%}$ & $79.50\%$ & $84.20\%$ & $\pmb{84.56\%}$ & $\pmb{85.62\%}$ & $85.05\%$ & $\pmb{84.60\%}$ & $83.59\%$\\
    \textbf{$k$NN} & $\pmb{86.95\%}$ & $86.51\%$ & $\pmb{85.62\%}$ & $84.64\%$ & $\pmb{82.33\%}$ & $82.05\%$ & $\pmb{81.08\%}$ & $79.25\%$ & $85.29\%$ & $\pmb{85.58\%}$\\
    \textbf{Drop-200} & $64.63\%$ & $\pmb{64.87\%}$ & $\pmb{79.09\%}$ & $77.67\%$ & $\pmb{73.22\%}$ & $72.77\%$ & $\pmb{74.51\%}$ & $73.91\%$ & $\pmb{75.65\%}$ & $74.11\%$\\
    \hline
\end{tabular}
    \vspace{-5pt}
\end{table*}

\section{Additional Experimental Results}

In this section, we present additional quantitative and qualitative experimental results. We first show defense results on RS-CNN \cite{liu2019relation} to further demonstrate the effectiveness of IF-Defense. Additionally, we prove that IF-Defense can boost the classification accuracy significantly compared to previous defenses when facing sparse input point clouds. Then, we conduct more ablation studies on the hyper-parameters $\lambda$, the contribution of SOR pre-processor, cross dataset evaluation of IF-Defense and the inference speed of different defense methods. We also present more visualizations of the defense points clouds to compare different defense methods qualitatively.

\subsection{Defense Results on RS-CNN}

Table \ref{rscnn_all_defense} illustrates the classification accuracy of RS-CNN under various attack and defense methods on the ModelNet40 dataset. Similarly, the optimization based IF-Defense with ConvONet achieves the best performance against most of the attacks, and the greatest improvement still comes from point dropping attack because it distorts object geometries. Interestingly, although RS-CNN also relies on neighboring points to extract features as DGCNN and PointConv, DUP-Net performs much better in this case. We hypothesize that this is because RS-CNN explicitly leverages the Euclidean distance between points as input, and thus can model the local point distributions. Consequently, RS-CNN is less sensitive to point distribution changes. 
Such property can guide our future design of point cloud networks.

\subsection{Defense Results on Sparse Point Clouds}

Though point cloud networks are shown to be robust against varying number of points in the input \cite{qi2017pointnet}, dropping more than 50\% of points still harm the performance seriously. Therefore, we study the results of different defense methods on sparse point clouds. We randomly drop 600 and 800 points from the original $1024$ points as input, and report the accuracy after defense in Table \ref{rand_600_defense} and Table \ref{rand_800_defense}, respectively. We observe significant improvement when applying IF-Defense. Because of the strong reconstruction capacity of implicit function networks, we can recover a correct object shape using only 200 input points, preserving its semantic information. Besides, we discover that PointNet is the most robust network among five victim models. This is because PointNet extracts per-point features individually and does not rely on interactions between local points.

% speed
\begin{table*}[tb]
    \caption{Comparison of inference time of different defense methods. The experiments were conducted on a Nvidia RTX 2080Ti GPU.}
    \label{speed}
    \centering
 \vspace{5pt}
\renewcommand\arraystretch{1.2}
\begin{tabular}{|c|c|c|c|c|c|c|c|}
    \hline
    \textbf{Defense} & \textbf{SRS} & \textbf{SOR} & \textbf{SOR-AE} & \textbf{DUP-Net} & \textbf{Ours-Mesh$^\dagger$} & \textbf{Ours-Opt$^\dagger$} & \textbf{Ours-Opt$^\ddagger$}\\
    \hline
    Time (second) & $1.1\times10^{-4}$ & $1.9\times10^{-3}$ & $0.05$ & $0.38$ & $0.49$ & $1.6$ & $1.0$\\
    \hline
\end{tabular}
    \vspace{-5pt}
\end{table*}

\subsection{More Ablation Studies}
\label{more_ablation}

\textbf{Distribution-aware loss weight.} We examine the effects of hyper-parameter $\lambda$ more extensively by testing against more attacks and employing another implicit function network ONet. Figure \ref{ablation_conv_appendix} shows the results for our optimization based IF-Defense with ConvONet. For most of the attacks, we observe that the accuracy improves as we increase $\lambda$ until reaching its optimum at $\lambda=500$. For larger $\lambda$, the accuracy becomes worse again. We draw similar observations when adopting ONet as the implicit function network from Figure \ref{ablation_onet_appendix}, where $\lambda=500$ is the optimal value for most of the attacks on all four victim models.

In conclusion, the selection of an appropriate balancing weight $\lambda$ is critical for our defense method, and we select an optimal $\lambda=500$ to balance the importance between accurate shape surfaces and uniform point distributions.

\textbf{SOR pre-processor.} We study the defense performance of IF-Defense with or without SOR pre-processor, where the optimization based IF-Defense with ConvONet is evaluated. From Table \ref{sor_ablation}, we conclude that SOR is effective for outlier removal because IF-Defense with SOR pre-processor achieves better classification accuracy against point perturbation, adding and $k$NN attacks which introduce many outliers. Also, the performance gain on $k$NN attack is the smallest among the first three attacks, indicating that $k$NN attack mainly causes on-surface perturbation. In contrast, since point dropping attack only removes salient points without outlier generation, SOR pre-processor even degrades the performance slightly. Overall, we prove that the success of IF-Defense mainly stems from implicit function network's ability to recover natural shapes and the learned uniform point distribution introduced by the optimization process. SOR only serves as a pre-processor to remove obvious outliers.

\textbf{Cross dataset evaluation.} In real-world scenarios, the distribution of the attacked point clouds may be different from that of the data used for training. Therefore, we conduct cross dataset evaluation to verify the generality of IF-Defense. We trained a ConvONet only on ShapeNet and test its performance against several attacks on ModelNet40. Compared with training on two datasets, the accuracy degradation is less than $2\%$ in all the cases. This shows the strong generalization ability of IF-Defense.

\textbf{Speed.} Algorithmic efficiency is also very important in real-world deployment. We compare the inference speed of different defense methods as shown in Table \ref{speed}. Because of the optimization process, two variants of the optimization based IF-Defense are more time-consuming than previous methods. Still, employing ConvONet reduces running time by $37.5\%$ compared to using ONet, indicating that IF-Defense can be accelerated when combined with more efficient implicit function networks. Also, batch processing can be utilized in offline applications to further improve the defense efficiency.

% visualization
\begin{figure*}[tb]
	\centering
	\includegraphics[width=1.0\textwidth]{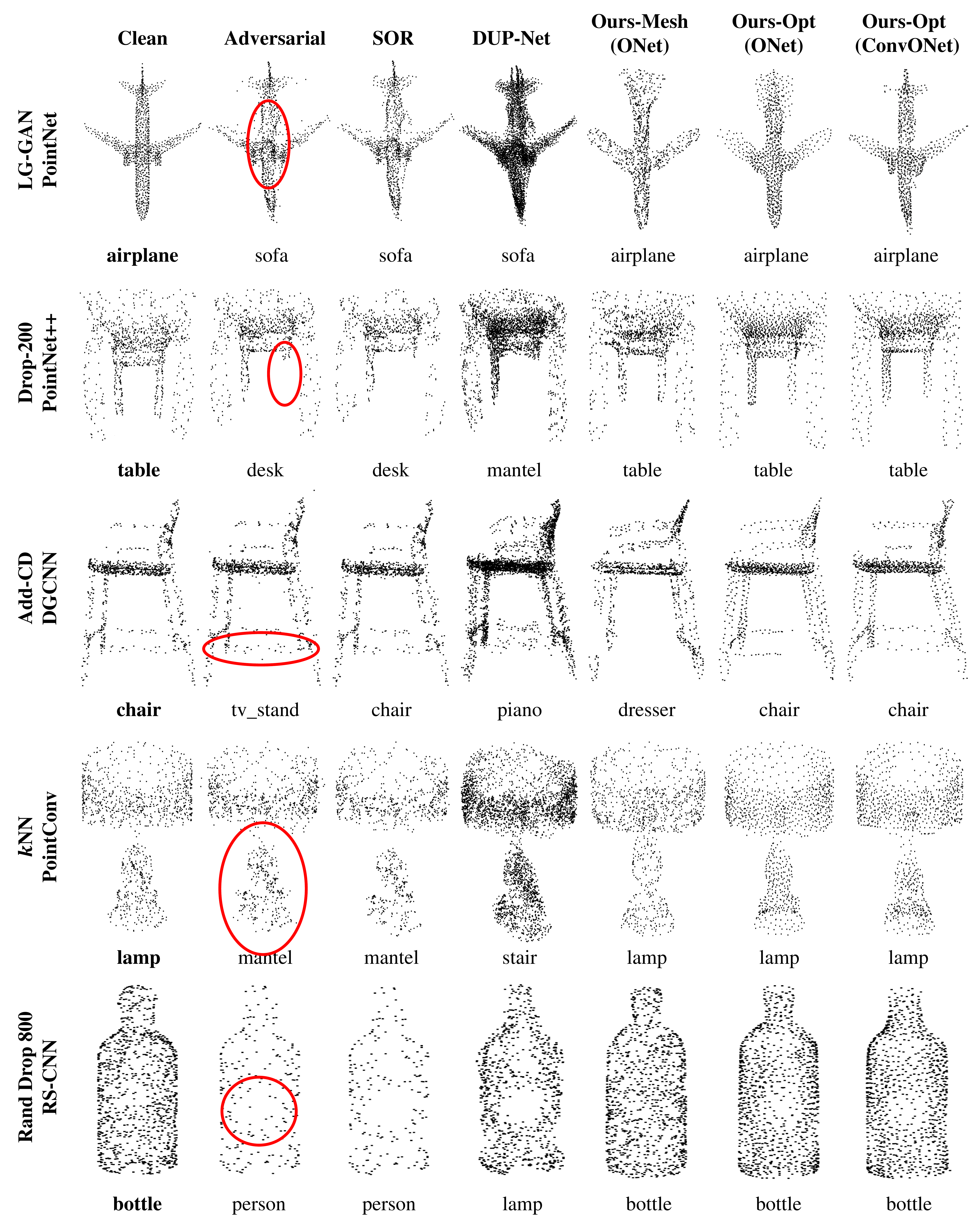}

	\caption{Visualizations of different defense results. The labels under each point cloud are the prediction outputs of the victim models.}
	\label{vis_defense_appendix}
    \vspace{-5pt}
\end{figure*}

\subsection{More Visualization Results}

We present more visualization results in Figure \ref{vis_defense_appendix} to compare the defense outputs of different methods against various attacks. For LG-GAN attack on PointNet, all three variants of IF-Defense successfully recover the straight body of the airplane. For point dropping attack on PointNet++, they re-introduce the lost leg of the table. In contrast, neither SOR nor DUP-Net can achieve such restorations. For point adding attack under Chamfer distance constraint on DGCNN, the re-meshing based IF-Defense fails to reconstruct the entire legs and their connection parts, whereas two optimization based IF-Defense succeed. For $k$NN attack on PointConv, DUP-Net outputs a lamp with a deformed base and denser points comparing to the clean one, thus being classified as a stair by the victim model. On the contrary, our methods restore point clouds with either original or uniform distributions. The last row shows the results of sparse point cloud on RS-CNN, where 800 points are randomly dropped from the original $1024$ points. Neither SOR nor DUP-Net can complete the large holes on the object surface. Besides, because of the strong representation capacity of implicit function networks, the output point clouds of all three variants of IF-Defense clearly represent a bottle, thus being correctly classified by RS-CNN.

\end{document}